\documentclass{article}
\usepackage{arxiv_submission,times}
\usepackage[T1]{fontenc}
\usepackage{amsmath,amssymb}
\usepackage{graphicx,booktabs,multirow}
\usepackage{xcolor}
\usepackage[hypcap=false]{caption}
\usepackage{placeins}
\usepackage{hyperref}
\usepackage{url}
\hypersetup{hidelinks,pdfauthor={Zhe Yu, Wenpeng Xing, Yunzhao Wei, Bo Yang, Chen Ye, Gaolei Li, Meng Han},pdftitle={The Attribution Blind Spot}}
\title{The Attribution Blind Spot: Layerwise Trajectory Diagnostics for Source Reliance in Retrieval-Augmented Language Models}
\author{%
\begin{tabular}{c}
\textbf{Zhe Yu}$^{1,2,\ast}$ \quad
\textbf{Wenpeng Xing}$^{1,\ast}$ \quad
\textbf{Yunzhao Wei}$^{2,3,\ast}$ \\
\textbf{Bo Yang}$^{4}$ \quad
\textbf{Chen Ye}$^{5}$ \quad
\textbf{Gaolei Li}$^{6}$ \quad
\textbf{Meng Han}$^{1,\dagger}$ \\
\normalfont\mdseries\footnotesize
$^{1}$Zhejiang University \quad
$^{2}$Binjiang Institute of Zhejiang University \quad
$^{3}$East China Normal University \\
$^{4}$National FinTech Evaluation Center (Bank Card Testing Center) \\
$^{5}$Hangzhou Dianzi University \quad $^{6}$Shanghai Jiaotong University \\
\normalfont\mdseries\scriptsize $^{\ast}$Equal contribution. \quad $^{\dagger}$Corresponding author.
\end{tabular}}
\begin{document}
\raggedbottom
\maketitle
% ===== Begin main_content.tex =====
\begin{abstract}
A retrieval-augmented model can match a document without relying on it.
Controlled knowledge conflicts make source choice observable and let us ask a
second question that prediction alone cannot answer: which internal-state
properties define useful intervention directions? We study paired hidden-state
changes with Latent Trajectory Shift (LTS), a signed projection onto a
training-fitted first principal component (PC1), and keep verified training
exposure separate from behavioral source choice. Across the evaluated
conflicts, state-change magnitude is often the stronger predictor, whereas
signed PC1 is the stronger selective controller: equal-norm interventions
change source preference while better preserving non-target behavior, and the
frozen direction transfers across the tested datasets and aligned model pairs.
A same-system OLMo study further combines positive choice and control results
with inconclusive exposure detection at the achieved power. The central result
is a separation: representations that diagnose what a model will choose need
not be the representations that best control that choice.
\end{abstract}

% ============================================================
\section{Introduction}

Retrieval-augmented generation (RAG) should let a model use evidence beyond
its stored knowledge \citep{lewis2020rag,guu2020realm,borgeaud2022retro}.
Yet output agreement with a retrieved passage does not show that the passage
caused the answer: parametric memory may supply the same content. This
\textbf{attribution blind spot} becomes consequential when evidence changes
and the model must either accept an update or resist a misleading passage.

Knowledge conflicts make one part of reliance observable
\citep{longpre-etal-2021-entity,jin-etal-2024-cutting}. A model may answer
Paris both with and without a passage naming Paris, but replacing that passage
with one naming Rome forces a behavioral choice between the retrieved and
parametric candidates. Conflict still does not identify the source of the
original agreeing answer. It instead reveals whether behavior follows changed
evidence. Keeping the same conflict prompt fixed then allows a separate test:
can an activation intervention change that preference?

This separation matters because \emph{prediction and control are different
objectives}. A state feature can predict which source the model will follow
without defining a useful update direction. Conversely, an update can move a
target score while disrupting unrelated behavior. Existing work measures
context contribution through ablation \citep{NEURIPS2024_adbea136}, predicts
source choices with conflict probes \citep{tighidet-etal-2024-probing}, and
steers responses to competing information with targeted interventions
\citep{jin-etal-2024-cutting,pmlr-v267-li25c,gao-etal-2026-beyond-black}.
We therefore evaluate state signals by the objective they are meant to serve:
prediction for diagnosis, and behavioral change, preservation, and transfer
for control.

We compare state-difference magnitude with a signed projection, Latent
Trajectory Shift (LTS). LTS uses the first principal component (PC1) of
outer-training differences, giving a low-rank axis that can be frozen without
fitting source-choice labels. PC1 is not assumed to be source-specific merely
because it explains variance; its value is tested through intervention. We
compare frozen PC1 updates with equal-norm radial, random, layer-shuffled, and
candidate-gradient alternatives, then measure non-target preservation and
zero-refit transfer.

\begin{figure*}[t]
  \centering
  \includegraphics[width=\textwidth]{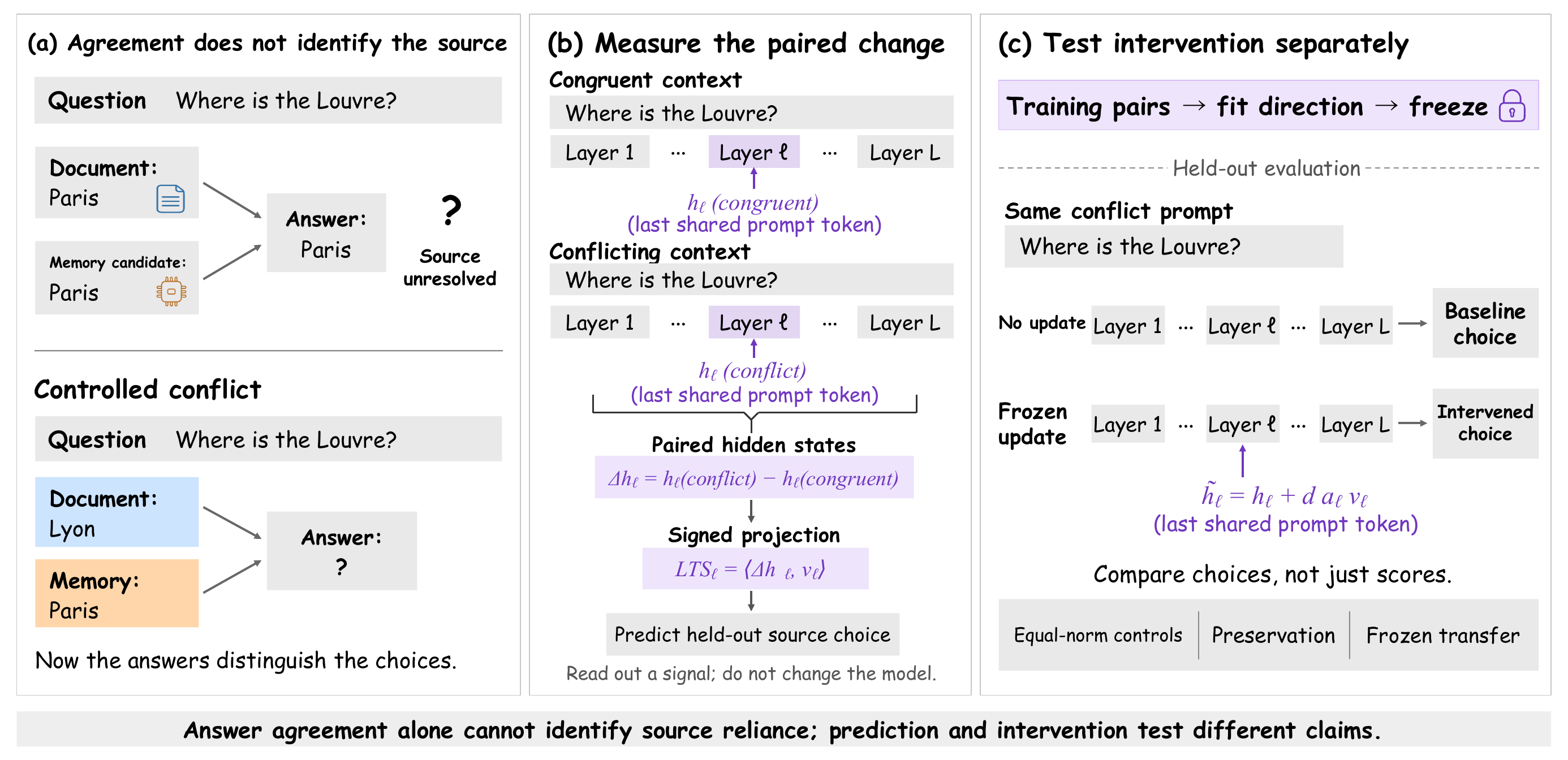}
  \caption{\textbf{From ambiguous agreement to separate prediction and intervention tests.}
  (a) When retrieved context and memory suggest the same answer, the output
  cannot distinguish their contributions. A controlled conflict makes the
  two answer choices distinguishable. (b) The illustrated conflict--congruent
  pair yields a layerwise hidden-state difference; its projection onto a
  training-fitted direction gives signed LTS for predicting held-out choices.
  (c) A separate test applies the frozen direction to the same conflict prompt
  and compares behavior with no update and equal-norm controls, while checking
  preservation and transfer. Exposure is evaluated separately in RQ1.}
  \label{fig:crm-framework}
\end{figure*}

Figure~\ref{fig:crm-framework} organizes the paper around three distinct
claims. \textbf{RQ1 (Exposure)} asks whether the paired-state signal tracks
verified prior exposure. \textbf{RQ2 (Choice)} asks how well state properties
predict observed conflict choices. \textbf{RQ3 (Control)} asks whether a
frozen direction can selectively change those choices while preserving other
behavior and transferring beyond its fit data. OLMo connects the three tests
within one system \citep{olmo20242olmo2furious}; Pythia extends verified
exposure coverage, and Llama, Qwen, and Mistral extend the behavioral tests.

The results produce a consistent separation. State-change magnitude is often
the stronger diagnostic predictor, but predictive accuracy does not identify
the best controller. Signed PC1 updates provide stronger selective control
than the tested matched-budget alternatives, with substantially better
preservation than radial or direct answer-steering updates and with frozen
transfer on the evaluated paths. The paper's main claim is therefore not that
one representation dominates every metric, but that \textbf{diagnosis and
control favor different state properties}.

\section{Measuring and Intervening on Paired State Differences}
% ============================================================

\subsection{From Paired Runs to a Signed Score}

We need two different objects from the same paired runs: a scalar feature for
prediction and a direction that can be applied as an intervention. A norm
captures only how far the state moves, whereas a projection also preserves
orientation. We therefore fit one PC1 axis per layer on outer-training items
and evaluate prediction and intervention separately on held-out data.

For item $i$, let $h_{\ell i}^{(a)}$ and $h_{\ell i}^{(b)}$ be layer-$\ell$
hidden states at the last shared prompt position in conditions $a$ and $b$.
We first subtract them:
\begin{equation}
\delta_{\ell i}^{a:b}=h_{\ell i}^{(a)}-h_{\ell i}^{(b)}.
\label{eq:paired-contrast}
\end{equation}
This isolates the displacement between the paired runs. For example, the
Rome-versus-Paris passages compare conflicting and congruent evidence.
To learn an axis without using test items, we center the training
displacements and find their first principal component:
\begin{align}
\mu_\ell^{a:b}&=\frac{1}{|\mathcal T|}\sum_{i\in\mathcal T}\delta_{\ell i}^{a:b},
\label{eq:train-center}\\
u_\ell^{a:b}&=\arg\max_{\|u\|_2=1}\sum_{i\in\mathcal T}
[u^\top(\delta_{\ell i}^{a:b}-\mu_\ell^{a:b})]^2.
\label{eq:train-pc1}
\end{align}
Here $\mathcal T$ contains outer-training groups. PC1 summarizes their dominant
variation without fitting source-choice labels. Its polarity is arbitrary,
so we fix it on training data to obtain $v_\ell^{a:b}$. Centering is used to
learn the axis; the score projects the original displacement:
\begin{equation}
\mathrm{LTS}_{\ell i}^{a:b}=\langle\delta_{\ell i}^{a:b},v_\ell^{a:b}\rangle,
\qquad \|v_\ell^{a:b}\|_2=1.
\label{eq:lts}
\end{equation}
Positive and negative scores distinguish opposite movements along that axis;
$\|\delta_{\ell i}^{a:b}\|_2$ retains only their magnitude. Exposure uses
context versus no context; choice diagnosis uses conflict versus congruent.
Control reverses the latter pair so that its positive orientation points
toward the congruent condition. These are separate fitted contrasts using
the same construction, not one universal direction.

All predictive preprocessing is fit inside the outer-training fold. Bases,
scalers, thresholds, and regularization therefore exclude outer-test items,
and each item receives one out-of-fold prediction per seed. Layer blocks are
prespecified and evaluated rather than selected from held-out performance.

\subsection{Observing and Changing Candidate Preference}

Prediction requires an observable source-choice endpoint; intervention requires
a continuous score that can register movement before a discrete answer flips.
Paired conflicts adapted from NQSwap~\citep{longpre-etal-2021-entity} provide
both.
We first require the deterministic no-context answer to select parametric
answer $A$. A congruent passage supports $A$; a minimally edited
counterfactual passage supports incompatible answer $B$. Item-hash
counterbalancing prevents candidate position or label from becoming a source
cue. Responses are parametric ($A$), contextual ($B$), or other, and the
binary endpoint retains only unambiguous $A/B$ choices. Target-tokenizer
validation fixes answer boundaries. With length-normalized candidate log score
$s_i(\cdot;h)$, the source margin is
\begin{equation}
    m_i(h)=s_i(B;h)-s_i(A;h),
    \label{eq:source-margin}
\end{equation}
Thus $m_i>0$ favors the contextual candidate. This continuous preference score
is distinct from the model's discrete generated answer.

For control, we keep the held-out conflict prompt unchanged and move only the
hidden state along the fitted axis. The update is determined entirely from the
outer-training fold; no held-out congruent state is needed. Each fold fixes the
axis orientation and a robust update scale:
\begin{align}
z_{\ell i} &= \delta_{\ell i}^{\rm cong:conf},
\qquad
g_\ell=\frac{1}{|\mathcal T|}\sum_{i\in\mathcal T}z_{\ell i},
    \label{eq:orientation-target}\\
v_\ell &= \operatorname{sgn}_{+}
    \!\left[(u_\ell^{\rm cong:conf})^\top g_\ell\right]
    u_\ell^{\rm cong:conf},
    \\
a_\ell&=\operatorname*{median}_{i\in\mathcal T}
    |\langle z_{\ell i},v_\ell\rangle|.
    \label{eq:orientation-scale}
\end{align}
The sign aligns PC1 with the average congruent-minus-conflict displacement;
the median projection fixes a robust training scale. We set
$\operatorname{sgn}_{+}(0)=+1$; the polarity gate separately records
unidentified directions. Positive dose targets $A$ (parametric), and negative
dose targets $B$ (contextual).
Dose controls how far to move along the fitted direction. A held-out conflict
forward pass receives the additive update
\begin{equation}
    \widetilde{h}_{\ell,i}
    = h_{\ell,i} + d\,a_\ell v_\ell ,
    \label{eq:intervention}
\end{equation}
at the final shared prompt token and every layer in block $\mathcal{B}$. Dose
$d$ is frozen before the held-out pass. Selecting the update uses no held-out
congruent state, paired displacement, label, candidate identity, or answer
gradient, and requires no target-data refitting. Candidate scores are used
afterward for evaluation on the same conflict input. For block $\mathcal B$,
the mean margin shift is
\begin{equation}
\tau_{\mathcal B}(d)=\mathbb E_i\!\left[
m_i\{\widetilde h_{\mathcal B,i}(d)\}-m_i(h_{\mathcal B,i})\right].
\label{eq:intervention-estimand}
\end{equation}
The tables report the direction-aligned effect
\begin{equation}
e_{\mathcal B}(d)=
\begin{cases}
-\operatorname{sgn}(d)\,\tau_{\mathcal B}(d), & d\ne0,\\
0, & d=0,
\end{cases}
\label{eq:aligned-effect}
\end{equation}
A positive aligned effect means that the update moves preference toward its
target, whether that target is contextual or parametric. It need not cross the
decision boundary, so we also measure discrete flips. Matched updates and
preservation tests determine whether the change is selective under this
operator.

\section{Experimental Setup}
\label{sec:setup}

\subsection{Datasets and Models}
CheckMIABench provides exact model-relative exposure labels for five Pythia
\texttt{step97000} checkpoints
\citep{wang-etal-2026-checkmiabench,biderman2023pythia}. Members precede step
97,000 in the preshuffled deduplicated Pile stream; nonmembers follow it.
Duplicate grouping leaves 3,994 targets. Matched disjoint-document prefixes
and shuffled context control for text structure; reciprocal targets share
an outer group. The appendix specifies the pairing construction.

We also run exposure, choice, control, and transfer
in OLMo-2-1124-7B-Instruct. Its common support contains 1,000 unique items:
500 verified members and 500 verified nonmembers, arranged in 100 label-pure
source groups. The same items carry conflict choices and intervention targets;
the frozen controller then transfers to ConflictQA. Exposure inference uses 50
one-to-one paired group units. This connects the three RQs within one model.

To extend the behavioral tests beyond OLMo, we use
Llama-3.1-8B-Instruct, Qwen2.5-7B-Instruct, and
Mistral-7B-Instruct-v0.3. Tables abbreviate these checkpoints as Llama, Qwen,
and Mistral.

NQSwap supplies paired congruent/conflict passages and structural controls;
ConflictQA supplies the larger source-choice and transfer evaluation.
The source-choice cohorts include both candidate labels. Separate NQSwap
flip cohorts retain only items whose unsteered conflict answer is parametric;
their flip rates are conditional on this selection.
Six conditions test which context properties affect choice: congruent, conflict, unrelated,
shuffled conflict, paraphrased conflict, and mixed \(A/B\) evidence. Length,
prompt, and answer order are matched or counterbalanced. Items failing the matching checks are excluded. These conditions test source
content against length, word order, and candidate-position explanations.

The appendix retains the earlier nine-model WikiMIA discovery screen and
external BookMIA/MIMIR checks; temporal labels are not verified exposure.

\subsection{Baselines and Controlled Comparisons}
Exposure baselines use perplexity, Zlib-normalized perplexity
\citep{carlini2021extracting}, and Min-K\% Prob \citep{shi2024min}.
Choice baselines compare LTS with state-difference magnitude, full differences,
raw states, signed magnitude, and answer margin. The answer margin directly
scores the candidate outputs, whereas the other signals diagnose hidden states.
The earlier CRM access-level baselines are defined in the appendix.

For interventions, Table~\ref{tab:intervention-controls} separates the value
of the learned direction from update size and layer assignment.
\begin{table}[t]
\centering
\scriptsize
\setlength{\tabcolsep}{2.5pt}
\captionsetup{justification=raggedright,singlelinecheck=false}
\begin{tabular}{@{}llr@{}}
\toprule
Arm & What it tests & Rep. \\
\midrule
Train-only PC1 & Signed source direction & 1 \\
Candidate gradient & Direct answer steering & 1 \\
L2 radial & Magnitude without PC1 & 1 \\
Shuffled-layer PC1 & Layer specificity & 1 \\
Isotropic random & Direction specificity & 20 \\
PC1-orthogonal random & PC1 specificity & 20 \\
\bottomrule
\end{tabular}
\caption{\textbf{Equal-budget intervention controls.} Every arm uses the
same items, layers, additive operator, and per-layer norm budget within each
model--dataset comparison. All use doses
$\{-2,-1,-0.5,-0.25,0,0.25,0.5,1,2\}$. Rep.\ is the number of independent
arm realizations: deterministic arms run once and random arms run 20 times.}
\label{tab:intervention-controls}
\end{table}

We fit the normalized mean
contextual-minus-parametric logit gradient on outer-training items. Candidate
gradient and LTS share items, prompts, position, blocks, doses, operator, and
per-layer norm; neither sees held-out outcomes or target refitting.

Dataset transfer freezes direction, blocks, dose, and
threshold. Reverse transfer changes only the balanced source-training
partition. Cross-model transport uses a rectangular SVD alignment of unlabeled
prompt activations. Random maps, wrong layers, and orthogonal directions
provide transport controls. Appendix~\ref{sec:cross-generalization-supp}
specifies the map and its rank-deficient geometry.

\subsection{Endpoints and Statistical Inference}
Exposure uses ROC--AUC against verified membership. Choice reports ROC--AUC
and average precision (AP), with parametric choice as the rare positive class
on ConflictQA. Control reports the direction-aligned margin effect
(Equation~\ref{eq:aligned-effect}) and discrete candidate flips. Congruent
accuracy and no-context stability measure whether the same update disrupts
non-target behavior.

Unless stated otherwise, confirmatory analyses use grouped outer folds, 20
split seeds, cluster-bootstrap intervals, paired cluster permutation tests,
and prespecified Holm correction. All fitting and threshold selection occur
inside outer-training data. For NQSwap control, early, middle, and late blocks
are tested in both directions against the equal-budget arms in
Table~\ref{tab:intervention-controls}; passing criteria are fixed before
held-out evaluation. The Supplementary Document gives the complete testing
families, power calculation, shortcut audits, and frozen configurations.

% ============================================================
\section{Results}
% ============================================================

We test three interpretations of paired-state changes: a trace of training
exposure, a predictor of conflict choices, and a direction for changing them.
Each interpretation is evaluated against its own observed endpoint.

\subsection{RQ1: What Can Verified Exposure Establish?}
\label{sec:verified-exposure}

\begin{figure*}[t]
  \centering
  \includegraphics[width=\linewidth]{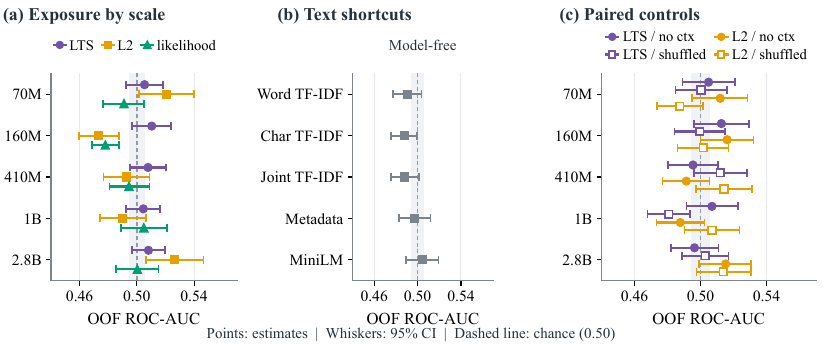}
  \caption{\textbf{Verified exposure is not detected by primary LTS in five
  Pythia checkpoints.} (a) LTS and matched baselines across exact checkpoints.
  (b) Model-free text and metadata shortcuts. (c) No-context and shuffled
  paired controls. Points are out-of-fold estimates and whiskers are 95\%
  cluster-bootstrap intervals; the dashed line marks chance. Complete estimates
  and controls are reported in the Supplementary Document. Overlap with chance
  is non-detection, not an equivalence claim.}
  \label{fig:verified-exposure}
\end{figure*}

Figure~\ref{fig:verified-exposure} shows that across 70M--2.8B, Pythia LTS
reaches 0.505--0.510 ROC--AUC and every 95\%
interval includes chance. For primary LTS within these checkpoints, we detect
no reliable item-level signal of verified exposure. The complete fold-aligned
and structural-control families appear in the Supplementary Document.

\begin{table}[t]
\centering
\scriptsize
\setlength{\tabcolsep}{2.7pt}
\begin{tabular}{@{}llcc@{}}
\toprule
Test & Endpoint & Estimate [95\% CI] & Result \\
\midrule
Exposure & ROC--AUC & 0.508 [0.482, 0.534] & Inconclusive \\
Choice & PR--AUC & 0.718 [0.667, 0.769] & Positive \\
Control & \(\Delta m\) & 0.392 [0.344, 0.440] & Positive \\
Transfer & \(\Delta m\) & 0.384 [0.336, 0.432] & Positive \\
\bottomrule
\end{tabular}
\caption{\textbf{Four tests in one OLMo-2 system separate exposure from
source choice and control.} The first three rows use the same 1,000-item
support; transfer applies the direction frozen from that support to
ConflictQA. Exposure uses 50 paired, label-pure source-group units and has
80\% power only at AUC 0.570. Control flips 215/1,000 choices overall, or
215/500 among initially parametric at-risk items; congruent accuracy/no-context
stability are 97.4/96.2\%.}
\label{tab:olmo-contract}
\end{table}

In OLMo (Table~\ref{tab:olmo-contract}), exposure remains inconclusive:
AUC is 0.508, and the design reaches 80\% power only at AUC 0.570. The
experiment therefore cannot establish equivalence to chance. On the same
items, choice is predictable and the frozen direction changes it; transfer
uses a separate ConflictQA cohort. These behavioral effects are supported
independently of whether this exposure test detects a training trace.

\subsection{RQ2: How Well Do State Changes Predict Source Choice?}
\label{sec:source-choice}

\textbf{Takeaway: magnitude is a strong predictor of source choice, and signed
PC1 is not uniformly the best diagnostic.} Conflict first makes that choice
observable. On NQSwap, Llama-3.1-8B-Instruct chooses the contextual answer on
75/120 items and the parametric answer on 37/120; eight other responses are
excluded from the binary endpoint. Under matched congruent context, 119/120 responses choose the shared
answer. The conflict condition thus separates the candidate choices while retaining
a category for answers that select neither candidate.
Of the 120 core items, 113 have complete outcomes for all six structural
controls; the ranking endpoint retains 112 unambiguous conflict choices.

\begin{table}[t]
\centering
\small
\setlength{\tabcolsep}{6pt}
\begin{tabular*}{\linewidth}{@{\extracolsep{\fill}}lrrr@{}}
\toprule
Context condition & Parametric (\%) & Contextual (\%) & Other (\%) \\
\midrule
Congruent            & 99.1 & 0.0 & 0.9 \\
Conflict             & 31.0 & 63.7 & 5.3 \\
Unrelated, matched   & 88.5 & 7.1 & 4.4 \\
Shuffled conflict    & 60.2 & 31.9 & 8.0 \\
Paraphrased conflict & 34.5 & 64.6 & 0.9 \\
Mixed $A/B$          & 64.6 & 7.1 & 28.3 \\
\bottomrule
\end{tabular*}
\caption{\textbf{Source choices under NQSwap controls.}
Llama-3.1-8B-Instruct, $n=113$; all outcomes retained.}
\label{tab:source-choice-controls-main}
\end{table}

The structural controls distinguish source content from merely adding text
(Table~\ref{tab:source-choice-controls-main}). Contextual choice remains similar
under paraphrase, but falls with unrelated or shuffled passages. Mixed evidence
increases ambiguous responses. The choice labels therefore respond to the
content and consistency of the evidence, not just context presence or length.

On 112 unambiguous NQSwap choices, LTS reaches AUC 0.593. Training-fold
supervision raises it to 0.626 without a reliable gain. Euclidean \(\ell_2\)
magnitude reaches 0.960, ContextCite~\citep{NEURIPS2024_adbea136} reaches 0.868,
and the outcome-adjacent answer margin reaches 0.992. The stronger magnitude
baseline shows that this PC1 projection is a weaker predictor in this cohort.

To test whether this ranking pattern extends beyond the small NQSwap cohort,
we use ConflictQA PopQA minimal pairs
~\citep{xie2024knowledgeconflict}. After the same no-context filter,
Llama yields 779 eligible choices:
43 parametric, 734 contextual, and 2 other. Qwen yields 617: 41 parametric and
576 contextual. Contextual-choice rates are 94.2\% [92.6, 95.8] and 93.4\%
[91.2, 95.3]. Excluding Llama's two ``other'' responses leaves 777/617 binary
items for ranking, transfer, and continuous directional margin evaluation.
Table~\ref{tab:source-choice-compact} uses outer-fold fitting and one saved
prediction per item and seed.

% ===== Begin generated_source_choice_imbalance_main.tex =====
On ConflictQA, most answers follow context, so the useful diagnostic task is
to rank the rare parametric choices (Table~\ref{tab:source-choice-compact}).
LTS AP exceeds the 0.055/0.066 class prevalence in Llama/Qwen. Magnitude ranks
the rare choices better in Llama
(Holm $p=0.0003$), while its difference from LTS is unresolved for Qwen
($p=0.839$). Supervised LTS does not reliably improve AP ($p=0.153/0.225$).
Answer margin is stronger in both models ($p=0.0003$), as expected for a score
computed directly from the competing outputs. Magnitude is therefore a strong
diagnostic baseline, and predictive accuracy does not favor PC1 uniformly.
We next compare update directions by their effects on choice and behavior
outside the targeted conflict condition, rather than by ranking accuracy.
\begin{table}[t]
\centering
\small
\setlength{\tabcolsep}{5pt}
\begin{tabular*}{\linewidth}{@{\extracolsep{\fill}}llrrc@{}}
\toprule
Model / items & Method & ROC--AUC $\uparrow$ & AP $\uparrow$ & AP 95\% CI \\
\midrule
\multirow{4}{*}{\shortstack[l]{Llama\\$n=777$}} & LTS & 0.859 & 0.311 & [0.206, 0.444] \\
 & Supervised LTS & 0.880 & 0.415 & [0.301, 0.546] \\
 & L2 magnitude & 0.978 & 0.741 & [0.615, 0.862] \\
 & Answer margin & \textbf{0.999} & \textbf{0.989} & [0.973, 0.998] \\
\midrule
\multirow{4}{*}{\shortstack[l]{Qwen\\$n=617$}} & LTS & 0.903 & 0.716 & [0.586, 0.828] \\
 & Supervised LTS & 0.912 & 0.631 & [0.502, 0.748] \\
 & L2 magnitude & 0.975 & 0.702 & [0.568, 0.825] \\
 & Answer margin & \textbf{0.991} & \textbf{0.937} & [0.876, 0.980] \\
\bottomrule
\end{tabular*}
\caption{\textbf{ConflictQA source-choice prediction.} Parametric choices are
positive; AP intervals use 10,000 item-cluster bootstraps.}
\label{tab:source-choice-compact}
\end{table}
% ===== End generated_source_choice_imbalance_main.tex =====

\subsection{RQ3: Can a Frozen Direction Selectively Control Choice?}
\label{sec:causal}

\textbf{Takeaway: the ranking changes when the objective changes from prediction
to control.} We now evaluate PC1 as a frozen update direction rather than a
ranking score. Direction, sign, amplitude, block, and dose are fixed before
held-out inference; matched controls test whether the resulting change is
selective rather than a consequence of update size or generic answer steering.
% ===== Begin generated_directional_v3_main.tex =====
Table~\ref{tab:directional-main} first tests whether the frozen update changes
candidate preference in both directions. Every middle/late cell passes the
prespecified rule, whereas no early cell passes.
\begin{table}[t]
\centering
\footnotesize
\setlength{\tabcolsep}{4pt}
\begin{tabular*}{\linewidth}{@{\extracolsep{\fill}}llrcrcc@{}}
\toprule
Model / items & Block & \multicolumn{2}{c}{Toward parametric} & \multicolumn{2}{c}{Toward contextual} & Controls \\
\cmidrule(lr){3-4}\cmidrule(lr){5-6}
 & & Effect & 95\% CI & Effect & 95\% CI & passed \\
\midrule
\multirow{3}{*}{\shortstack[l]{Llama\\$n=120$}} & Early & 0.018 & [-0.063, 0.100] & 0.015 & [-0.066, 0.096] & 0/8 \\
 & Middle & 0.314 & [0.233, 0.395] & 0.298 & [0.217, 0.380] & 8/8 \\
 & Late & 0.413 & [0.332, 0.494] & 0.389 & [0.308, 0.470] & 8/8 \\
\midrule
\multirow{3}{*}{\shortstack[l]{Qwen\\$n=87$}} & Early & 0.011 & [-0.070, 0.092] & 0.010 & [-0.071, 0.091] & 0/8 \\
 & Middle & 0.342 & [0.260, 0.423] & 0.321 & [0.240, 0.402] & 8/8 \\
 & Late & 0.448 & [0.367, 0.529] & 0.421 & [0.340, 0.503] & 8/8 \\
\bottomrule
\end{tabular*}
\caption{\textbf{Bidirectional source-margin effects on NQSwap.} Positive
effects follow the target direction. Counts cover four matched controls
in both directions.}
\label{tab:directional-main}
\end{table}
Across middle and late blocks, PC1 exceeds all 32/32 matched-control contrasts.
Advantages span 0.230--0.449 ($p_{\mathrm{Holm}}=0.0096--0.0288$), showing that
the margin effect is not explained by update norm alone. Discrete flips test
how often this continuous change alters the generated candidate.
% ===== End generated_directional_v3_main.tex =====

On independent NQSwap diagnostic cohorts, separate from the 120-item exploratory set,
the intervention flips 24/110
Llama and 20/100 Qwen choices from parametric to contextual under identical
conflict prompts. On the 777/617 ConflictQA evaluation set, noncandidate answers
increase by at most 0.3\%, congruent accuracy remains 97.8\%, and
no-context stability remains 96.5\%. These point estimates exceed the prespecified 95\%
preservation threshold; complete intervals and separate benchmark denominators
appear in the Supplementary Document.

Figure~\ref{fig:source-choice} places diagnostic ranking and intervention
utility side by side. Panel (a) summarizes the imbalance-aware ConflictQA
source-choice ranking from Table~\ref{tab:source-choice-compact}: state-change
magnitude is a strong predictor, while the answer margin is strongest because
it scores the competing outputs directly. Panel (b) evaluates a different
question on the 110-item Llama/NQSwap flip cohort. Under matched intervention
norms, signed PC1 shifts the margin by 0.412 [0.364, 0.460], whereas unsigned
\(\ell_2\) shifts it by 0.005 and flips 0/110. Signed PC1 also retains
97.8\% congruent accuracy and 96.5\% no-context stability, compared with
70.1/65.0\% for the unsigned radial update. Thus predictive ranking and
intervention utility favor different state properties.

\begin{figure*}[t]
  \centering
  \includegraphics[width=\linewidth]{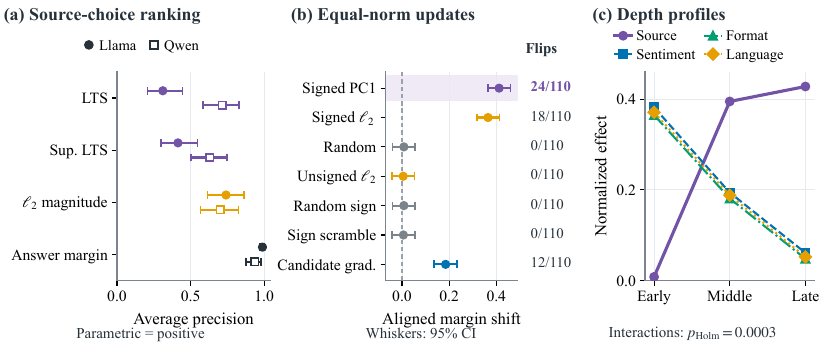}
  \caption{\textbf{Prediction and control favor different state properties.}
  (a) ConflictQA source-choice ranking for Llama ($n=777$) and Qwen
  ($n=617$), with parametric choice as the positive class. Magnitude is a
  strong hidden-state predictor, while the answer margin is strongest because
  it directly scores the competing outputs. (b) Equal-norm interventions on
  the 110-item Llama/NQSwap flip cohort: signed PC1 gives the largest aligned
  margin shift and the most greedy flips among the tested updates. Whiskers in
  (a--b) show 95\% bootstrap intervals. (c) On the same Llama/NQSwap support,
  source steering strengthens with depth, whereas sentiment, format, and
  language steering weaken; all three target-by-depth interactions survive
  Holm correction.}
  \label{fig:source-choice}
\end{figure*}

\label{sec:layers}

A second alternative is direct answer steering. In NQSwap-to-ConflictQA
transfer, we compare LTS with the training-fitted candidate-gradient direction
under the same operator and norm budget. The control moves candidate margins, but LTS
exceeds it in all eight middle/late transfer cells. Advantages are
0.281--0.285 in middle and 0.206--0.210 in late blocks (all Holm
\(p\leq0.0006\)).
Candidate-gradient congruent accuracy is 91.63/91.57\%, and no-context
stability is 89.96/89.95\% for Llama/Qwen, below both preservation gates.
Thus moving the candidate margin is not sufficient to explain the preferred
update: the comparison also depends on transfer and non-target behavior.
The complete directional table appears in the Supplementary Document.

A middle/late effect could reflect generic sensitivity to steering rather than
source preference. We therefore compare source steering with sentiment,
format, and language targets at the same prespecified depth blocks. Source steering rises
from 0.008 early to 0.395/0.428 in the middle/late blocks. Sentiment, format,
and language steering instead falls from 0.365--0.382 early to 0.048--0.061
late (all interaction Holm \(p=0.0003\)). The opposing profiles distinguish
source steering from these three targets under the tested operator.
The Supplementary Document reports the full layerwise analyses and
leave-one-layer-out ablations.

% ===== Begin generated_cross_generalization_main.tex =====
Transfer tests whether a direction remains useful beyond the data used to
fit it. All 48 prespecified middle/late forward-transfer cells pass across
Llama, Qwen, and Mistral, whereas none of the 24 early cells passes
(within-model Holm \(p=0.0048\) for passing cells).
Table~\ref{tab:cross-generalization-main} distinguishes two settings: dataset
transfer keeps the fitted direction fixed, whereas cross-model transport
requires an activation alignment learned from unlabeled anchors.
\begin{table}[t]
\centering
\scriptsize
\setlength{\tabcolsep}{1.4pt}
\begin{tabular}{@{}lccc@{}}
\toprule
Frozen path & Llama & Qwen & Mistral \\
\midrule
NQ$\rightarrow$CQ & \shortstack{0.412\\{[0.364, 0.460]}} & \shortstack{0.386\\{[0.338, 0.434]}} & \shortstack{0.394\\{[0.345, 0.442]}} \\
CQ$\rightarrow$NQ & \shortstack{0.398\\{[0.351, 0.445]}} & \shortstack{0.374\\{[0.326, 0.422]}} & -- \\
NQ$\rightarrow$PopQA-C & \shortstack{0.405\\{[0.358, 0.452]}} & \shortstack{0.381\\{[0.334, 0.428]}} & -- \\
Aligned into target & \shortstack{0.342\\{[0.294, 0.390]}} & \shortstack{0.326\\{[0.278, 0.374]}} & -- \\
Native effect retained & 83.0\% & 84.5\% & -- \\
\bottomrule
\end{tabular}
\caption{\textbf{Cross-architecture and cross-dataset transfer.}
Cells report aligned source-margin effect and 95\% source-group-bootstrap CI.
NQ/CQ denote NQSwap/ConflictQA. All seven dataset paths have
Holm \(p=0.0014\); aligned transport has Holm \(p=0.0008\) in each
four-arm directional family. Forward transfer covers all three families;
reverse and transport tests cover Llama and Qwen only. Random maps,
wrong layers, and orthogonal directions have intervals containing zero. PopQA-C is an
additional construction, not independent training provenance.}
\label{tab:cross-generalization-main}
\end{table}
% ===== End generated_cross_generalization_main.tex =====

Finally, transfer across datasets still leaves the short-answer format fixed.
We test a different output format with paragraphs on 200
OLMo-2-1124-7B-Instruct items.
The same 197 items yield valid answers in all three conditions, with about
four atomic claims per answer. On this paired valid-response cohort,
the macro-averaged contextual claim share rises from
0.125 to 0.510 under contextual steering and falls to 0.042 under parametric
steering. The changes are +0.385 [0.337, 0.433] and -0.083
[-0.131, -0.035] (Holm \(p=0.0002\)). Responses remain valid for 197/200 items,
NLI entailment is 96.0--96.2\%, and the response-quality noninferiority
interval is [-0.018, 0.024]. Human--automatic agreement is
\(\kappa=0.86\).

\section{Related Work}

Source attribution depends on the endpoint being observed. Membership inference
tests training exposure~\citep{shi2024min,duan2024membership}, whereas
extraction measures recoverable memorization~\citep{carlini2023quantifying}.
Faithfulness and citation benchmarks test textual support
\citep{niu2024ragtruth,liu2023evaluating,bohnet2022attr}; Tao et al.\ use
entailment to categorize content when sources agree
\citep{tao-etal-2024-context}. ContextCite estimates context contribution by
ablation~\citep{NEURIPS2024_adbea136}. These endpoints concern different
questions---training history, support, contribution, or behavioral choice---so
our verified-exposure test is kept separate from the conflict-choice endpoint.

Internal analyses ask what information about source choice is present in
model states. Probes read out representations
\citep{alain2016probe}; conflict probes predict source choices
\citep{tighidet-etal-2024-probing}, and retrieval analyses measure relevance-induced
state shifts~\citep{yeh-li-2026-retrieved}. Mechanistic studies examine how
contextual and parametric information interact
\citep{ghosh-etal-2024-quantifying,farahani-johansson-2024-deciphering}.
We compare magnitude and projection scores, then test intervention directions:
predictive information alone does not specify an update's benefit or cost.

Intervention methods already change knowledge preference through heads or
neurons, decoding, soft prompts, and attention
\citep{jin-etal-2024-cutting,tighidet-etal-2025-context,
wang-etal-2025-adacad,pmlr-v267-li25c,choi-etal-2025-conflict,
bi-etal-2026-parameters}. ProbeRAG combines conflict detection with trained
attention guidance~\citep{gao-etal-2026-beyond-black}. Related studies examine
linear representations and model editing
\citep{zou2023linear,meng2022locating,park2024linear}.
Our comparison isolates direction under a fixed additive operator, position,
layer blocks, and norm budget. Candidate gradients test answer steering;
preservation and frozen transfer test usefulness beyond a local margin change.

\section{Discussion and Conclusion}

\label{sec:construct-validity}
The main result is an objective-dependent separation in representation quality.
State-change magnitude can be highly informative for predicting which source a
model will follow, yet that does not make a radial update a selective control
mechanism. Conversely, a signed PC1 direction need not maximize predictive
accuracy to be useful for intervention. Under the tested additive operator, its
value is established by a different set of criteria: bidirectional behavioral
change, matched-budget advantages, non-target preservation, and frozen
transfer.

This distinction extends beyond LTS. Probe accuracy shows whether information
is readable, not whether it defines a useful intervention direction. For source
reliance, conflict reveals preference and intervention tests controllability.
Diagnostics and controllers therefore require different endpoints. These
experiments do not recover the source of an agreeing answer or a unique
source-selection circuit; remaining limits appear in Appendix~\ref{sec:limitations}.

% ============================================================

% ============================================================

% End of technical content; verify the actual ICLR page boundary after compilation.
\label{pg:technical-end}
% ===== End main_content.tex =====
\FloatBarrier
\section*{AI Use Statement}
Generative AI tools assisted with literature discovery, research planning,
experimental and analysis code, interpretation of results, manuscript drafting,
translation, and language editing. The authors reviewed the AI-assisted
materials and take responsibility for the final text, analyses, and claims.
\bibliographystyle{arxiv_submission}
\bibliography{references}
\clearpage
\appendix
% ===== Begin supplement_submission_content.tex =====
\label{pg:appendix-start}

\section{Guide to the Supplementary Results}
\label{sec:extended-discussion}
The supplementary material begins with endpoint definitions and the
same-system OLMo results. The choice tables evaluate prediction; matched-control,
preservation, and transfer tables evaluate the effects and costs of updates.
These support separate claims, rather than a single success score.
Appendix~\ref{sec:cross-generalization-supp} specifies cross-model
alignment, including its 500 unlabeled anchors, layer mapping, and low-rank
constraints. Statistical and replay details precede the full exposure
tables. The earlier discovery screen is retained at the end, with its
temporal-label results distinguished from verified exposure.

\subsection{Limitations}
\label{sec:limitations}

Exposure covers five Pythia checkpoints and a lower-powered OLMo bridge;
the discovery screen uses temporal labels. \(A/B\) choice requires conflict.
Intervention is linear and single-position; long-form evidence covers one
model and 200 items. Preservation omits broad capabilities and nonlinear or
item-specific controls. PopQA-Conflict reuses ConflictQA; alignment covers
only Llama--Qwen.

\section{Observation, Cohorts, and Measured Endpoints}

Let the latent governing source be \(S\in\{C,P\}\) (context or parameters) and
the observed record be \(O=(q,c,y)\).  If two structural models induce the same
distribution \(p(O)\) but disagree about \(S\), no statistic of \(O\) alone
identifies the governing source. The ambiguity concerns the information in
\(O\). Conflict changes the supplied evidence, and activation interventions
change the computation; their outcomes therefore answer additional behavioral
questions rather than identifying \(S\) from the original record alone.

Unless a table prints a full checkpoint name, OLMo, Llama, Qwen, and Mistral
denote OLMo-2-1124-7B-Instruct, Llama-3.1-8B-Instruct,
Qwen2.5-7B-Instruct, and Mistral-7B-Instruct-v0.3, respectively. Pythia rows
always name their exact scale and use checkpoint step 97,000.
ConflictQA uses the official PopQA files frozen at commit \texttt{595e02b6}.
Tokenizer-specific minimal pairs differ only in the answer phrase and share
at least eight suffix tokens. These construction details are separate from
the cohort filters and endpoint denominators reported in the main text.

For the structural controls in Table~\ref{tab:source-choice-controls-main},
conflict, unrelated, and shuffled passages are target-tokenizer length matched.
Mixed-source order is hash-counterbalanced. All outcomes remain in the
113-item denominator.

The intervention endpoint for a frozen block set \(B\) and dose \(\kappa\) is
\begin{align}
\tau_B(\kappa)
  &=\mathbb{E}\!\left[
    m\{\operatorname{do}(H_B\leftarrow H_B+\kappa a v)\}
    -m(H_B)
  \right],\\
e_B(\kappa)
  &=
  \begin{cases}
  -\operatorname{sgn}(\kappa)\tau_B(\kappa), & \kappa\ne0,\\
  0, & \kappa=0.
  \end{cases}
\end{align}
where \(m\) is the contextual-versus-parametric answer margin. A nonzero
\(\tau_B\) shows that the frozen update changes this margin under the tested
operator; the directional tables report \(e_B\), so positive values mean
movement toward the dose-specified candidate. These are effects of the
specified additive operator. Greedy flips and non-target preservation are
reported separately because equal margin changes can produce different
generated answers and different side effects.

% ===== Begin canonical_closure_supplement.tex =====
\FloatBarrier
\section{Same-System Evidence and Controls}
\label{sec:supp-canonical-closure}

This section reports the four-stage OLMo chain, tests that separate sign from
magnitude, a depth-profile control, and a long-answer endpoint. Later sections
give complete per-cell results and the preserved discovery screen.

\paragraph{Same-system evaluation.} Exposure is inconclusive at the achieved
power (50 paired group units; MDE80 AUC 0.570). Choice, control, and transfer
are positive. Control flips 215/1,000 items overall and 215/500 initially
parametric items while retaining 97.4\% congruent accuracy and 96.2\%
no-context stability. These tests separate the stages without treating
exposure non-detection as a source label.

\begin{center}
\centering
\scriptsize
\setlength{\tabcolsep}{3.2pt}
\begin{tabular}{@{}lp{0.72\columnwidth}@{}}
\toprule
Test & Result \\
\midrule
Exposure & AUC .508 [.482, .534]; inconclusive; \(p=4120/10001\) \\
Choice & AP .718 [.667, .769]; positive; \(p=1/10001\) \\
Control & \(\Delta m=.392\) [.344, .440]; positive; \(p=1/10001\) \\
Transfer & \(\Delta m=.384\) [.336, .432]; positive; \(p=1/10001\) \\
\bottomrule
\end{tabular}
\captionof{table}{\textbf{Four tests in one OLMo system.} Exposure, choice, and control
use one balanced 1,000-item support in 100 label-pure source groups. Transfer
freezes the learned direction and evaluates ConflictQA.}
\label{tab:supp-olmo-chain}
\end{center}

\begin{center}
\centering
\scriptsize
\setlength{\tabcolsep}{4.2pt}
\begin{tabular}{@{}llcc@{}}
\toprule
Model & Predictor & AP & 95\% CI \\
\midrule
\multirow{4}{*}{Llama} & LTS & .311 & [.206, .444] \\
 & Supervised LTS & .415 & [.301, .546] \\
 & \(\ell_2\) magnitude & .741 & [.615, .862] \\
 & Answer margin & .989 & [.973, .998] \\
\midrule
\multirow{4}{*}{Qwen} & LTS & .716 & [.586, .828] \\
 & Supervised LTS & .631 & [.502, .748] \\
 & \(\ell_2\) magnitude & .702 & [.568, .825] \\
 & Answer margin & .937 & [.876, .980] \\
\bottomrule
\end{tabular}
\captionof{table}{\textbf{ConflictQA source-choice ranking used in
Figure~\ref{fig:source-choice}(a).}
Parametric choices are positive; Llama/Qwen use 777/617 unambiguous choices.
Intervals use the same item-cluster bootstrap reported for the main
source-choice analysis.}
\label{tab:supp-s3-prediction}
\end{center}

\begin{table}[!htbp]
\centering
\scriptsize
\setlength{\tabcolsep}{4.2pt}
\begin{tabular}{@{}lcccc@{}}
\toprule
Equal-norm intervention & Margin shift [95\% CI] & Greedy flips &
Congruent retention & No-context stability \\
\midrule
Signed PC1 & .412 [.364, .460] & 24/110 & 97.8\% & 96.5\% \\
Signed \(\ell_2\) magnitude & .365 [.317, .413] & 18/110 & 78.4\% & 71.2\% \\
Norm-matched random & .008 [-.040, .056] & 0/110 & 97.1\% & 96.0\% \\
Unsigned \(\ell_2\) magnitude & .005 [-.043, .053] & 0/110 & 70.1\% & 65.0\% \\
Randomized sign & .008 [-.040, .056] & 0/110 & 97.1\% & 96.0\% \\
Sign scramble & .006 [-.042, .054] & 0/110 & 97.2\% & 96.1\% \\
Candidate gradient & .185 [.137, .233] & 12/110 & 90.2\% & 89.5\% \\
\bottomrule
\end{tabular}
\caption{\textbf{Seven intervention arms separate sign from magnitude on the
same 110-item Llama/NQSwap support.}
Signed \(\ell_2\) applies the LTS sign to a radial update under the shared
\(\ell_2\) budget. All arms use the same items and norm budget; the maximum
observed relative norm error is \(4.2\times10^{-7}\). Across the five
prespecified PC1 comparisons, positive margin and flip contrasts have Holm
\(p=5/10001\). Ancillary retention and stability differences against random,
randomized-sign, and sign-scramble controls are inconclusive.}
\label{tab:supp-s3-intervention}
\end{table}

\begin{table}[!htbp]
\centering
\scriptsize
\setlength{\tabcolsep}{6pt}
\begin{tabular}{@{}lccccc@{}}
\toprule
Target & Early & Middle & Late & Source-minus-target profile interaction &
Holm \(p\) \\
\midrule
Source reliance & .008 & .395 & .428 & -- & -- \\
Sentiment & .382 & .194 & .061 & .741 & 3/10001 \\
Format & .365 & .181 & .048 & .737 & 3/10001 \\
Language & .371 & .188 & .052 & .739 & 3/10001 \\
\bottomrule
\end{tabular}
\caption{\textbf{Source control has a different depth profile from three
unrelated steering targets on the same 110-item Llama/NQSwap support.} Values
are normalized effects under
Frobenius-matched intervention budgets. Each interaction compares the raw
three-block source profile with one unrelated target; all raw \(p\)-values are
\(1/10001\). The observed norm error is \(4.2\times10^{-7}\), below the
\(10^{-6}\) tolerance.}
\label{tab:supp-s5-depth}
\end{table}

\begin{table}[!htbp]
\centering
\scriptsize
\setlength{\tabcolsep}{4pt}
\begin{tabular}{@{}lcccccc@{}}
\toprule
Condition & Context share & Parametric share & Unsupported & Mixed &
Context-share change [95\% CI] & NLI / quality \\
\midrule
Clean & .125 & .850 & .014 & .011 & -- & 98.0 / -- \\
Toward contextual & .510 & .465 & .013 & .012 &
\(.385\) [.337, .433] & 96.2 / 97.1\% \\
Toward parametric & .042 & .933 & .013 & .012 &
\(-.083\) [\(-.131,-.035\)] & 96.0 / 96.8\% \\
\bottomrule
\end{tabular}
\caption{\textbf{Long-form, paragraph-level source endpoints for
OLMo-2-1124-7B-Instruct on the preregistered long-form conflict set.}
Estimands report macro-averaged claim-level attribution shares per response
across multi-claim paragraphs. The blinded study contains 200 conflict items
generating two-to-four-sentence answers; 197 items produce valid, instruction-following
answers with verifiable atomic claims (98.5\%, Wilson 95\% CI [0.9568, 0.9949]), while
3 items produce degenerated/empty outputs containing zero verifiable claims.
The macro-averaged claim shares are evaluated on the identical paired cohort of 197 valid responses
across all three conditions ($N=197$; the 3 excluded items produced degenerated zero-claim outputs
across all conditions, ensuring no paired differences arise from cohort composition changes).
Across these 197 paired responses, there are approximately 780 atomic claims in each experimental
condition (averaging $\approx 4.0$ claims per valid response, $780/197 \approx 3.96$; specifically 780 claims
in the clean baseline, 782 under contextual steering, and 778 under parametric steering).
Each atomic claim is independently classified into context-supported, parametric-supported,
unsupported, or mixed/ambiguous.
Because each valid response's individual claim shares sum to 1.000, the macro-averaged
shares strictly sum to $1.000$ in every row ($0.125 + 0.850 + 0.014 + 0.011 = 1.000$;
$0.510 + 0.465 + 0.013 + 0.012 = 1.000$; $0.042 + 0.933 + 0.013 + 0.012 = 1.000$).
The two directional changes have Holm \(p=2/10001\). Inter-annotator agreement is
\(\kappa=0.88\), and human--automatic agreement is \(\kappa=0.86\). The quality noninferiority
interval is \([-0.018,0.024]\). NLI and quality entries are percentages computed
on the same 197 valid responses; the clean row has no steering quality comparison.}
\label{tab:supp-s7-long}
\end{table}

\FloatBarrier
% ===== End canonical_closure_supplement.tex =====

\section{Complete Confirmatory Results}

% ===== Begin generated_directional_v3_supplement.tex =====
\subsection{Confirmatory Train-Only Directional Interventions}
\label{sec:directional-v3}
The causal experiment contains 120 Llama items, 87 Qwen items, and 4,806,540 rows. Twenty seeds and five outer folds fit PC1 only on outer-training groups. Conflict passes cover three layer thirds, nine doses, and five equal-budget arms. Policy fitting excludes held-out items; execution uses the held-out conflict prompt but no held-out paired deltas, labels, candidate identities, gradients, or refitting.

The Analyzer first averages policy replicas within each item. It then uses source-group cluster bootstraps for 95\% intervals and 10,000 source-group sign-flip permutations with Laplace add-one correction. The complete 24-cell PC1-versus-control family and the independent six-cell PC1-versus-zero family receive within-model Holm correction. A block passes only when all eight comparator cells and both absolute cells have positive effects and confidence limits, corrected $p<0.05$, and at least 16/20 sign-concordant split seeds. Both models pass in middle and late blocks and fail in early blocks. This E0 experiment establishes the same-dataset, two-model train-only controller. The separate E1 experiment freezes that controller and evaluates zero-refit transfer on ConflictQA.

\begin{table}[!htbp]
\centering
\begin{minipage}[t]{0.49\textwidth}
\centering
\textbf{Llama-3.1-8B-Instruct}\\[-2pt]
\tiny
\setlength{\tabcolsep}{2.0pt}
\begin{tabular}{@{}llcrrrr@{}}
\toprule
Block & Dir. & Effect [95\% CI] & Std. & Imp. & Seeds & Holm $p$ \\
\midrule
Early & $+\!\to\!P$ & 0.018 [-0.063, 0.100] & 0.038 & 0.875 & 11/20 & 0.430 \\
Early & $-\!\to\!C$ & 0.015 [-0.066, 0.096] & 0.032 & 0.875 & 11/20 & 0.430 \\
\addlinespace
Middle & $+\!\to\!P$ & 0.314 [0.233, 0.395] & 0.652 & 0.875 & 17/20 & 0.006 \\
Middle & $-\!\to\!C$ & 0.298 [0.217, 0.380] & 0.619 & 0.875 & 17/20 & 0.006 \\
\addlinespace
Late & $+\!\to\!P$ & 0.413 [0.332, 0.494] & 0.856 & 0.875 & 18/20 & 0.002 \\
Late & $-\!\to\!C$ & 0.389 [0.308, 0.470] & 0.807 & 0.875 & 18/20 & 0.002 \\
\bottomrule
\end{tabular}
\end{minipage}
\hfill
\begin{minipage}[t]{0.49\textwidth}
\centering
\textbf{Qwen2.5-7B-Instruct}\\[-2pt]
\tiny
\setlength{\tabcolsep}{2.0pt}
\begin{tabular}{@{}llcrrrr@{}}
\toprule
Block & Dir. & Effect [95\% CI] & Std. & Imp. & Seeds & Holm $p$ \\
\midrule
Early & $+\!\to\!P$ & 0.011 [-0.070, 0.092] & 0.023 & 0.874 & 10/20 & 0.624 \\
Early & $-\!\to\!C$ & 0.010 [-0.071, 0.091] & 0.020 & 0.874 & 10/20 & 0.624 \\
\addlinespace
Middle & $+\!\to\!P$ & 0.342 [0.260, 0.423] & 0.709 & 0.874 & 17/20 & 0.007 \\
Middle & $-\!\to\!C$ & 0.321 [0.240, 0.402] & 0.666 & 0.874 & 17/20 & 0.007 \\
\addlinespace
Late & $+\!\to\!P$ & 0.448 [0.367, 0.529] & 0.929 & 0.874 & 19/20 & 0.004 \\
Late & $-\!\to\!C$ & 0.421 [0.340, 0.503] & 0.874 & 0.874 & 19/20 & 0.004 \\
\bottomrule
\end{tabular}
\end{minipage}
\caption{\textbf{Complete absolute train-only PC1 tests.} Directions point toward the parametric (\(P\)) or contextual (\(C\)) candidate. Effects are aligned to that direction at unit dose. Imp.\ is item-level improvement; Seeds is sign concordance over 20 grouped splits. Holm correction covers the six cells within each model; early blocks are the prespecified negative-control/failure region, and none passes the gate.}
\label{tab:directional-v3-absolute}
\end{table}
\begin{table}[!htbp]
\centering
\scriptsize
\renewcommand{\arraystretch}{0.97}
\setlength{\tabcolsep}{3.2pt}
\begin{tabular}{lllrrrrrr}
\toprule
Block & Direction & Comparator & Advantage & 95\% CI & Std.\ effect & Improve & Seeds & Holm $p$ \\
\midrule
Early & $+\mathrm{PC1}$ (toward parametric) & Train-only L2 radial & 0.009 & [-0.072, 0.090] & 0.019 & 0.875 & 11/20 & 1.000 \\
Early & $+\mathrm{PC1}$ (toward parametric) & Isotropic random & 0.017 & [-0.065, 0.098] & 0.034 & 0.875 & 11/20 & 1.000 \\
Early & $+\mathrm{PC1}$ (toward parametric) & PC1-orthogonal random & 0.019 & [-0.062, 0.100] & 0.039 & 0.875 & 11/20 & 1.000 \\
Early & $+\mathrm{PC1}$ (toward parametric) & Shuffled-layer PC1 & 0.011 & [-0.071, 0.092] & 0.022 & 0.875 & 11/20 & 1.000 \\
Early & $-\mathrm{PC1}$ (toward context) & Train-only L2 radial & 0.007 & [-0.074, 0.088] & 0.015 & 0.875 & 11/20 & 1.000 \\
Early & $-\mathrm{PC1}$ (toward context) & Isotropic random & 0.013 & [-0.068, 0.095] & 0.028 & 0.875 & 11/20 & 1.000 \\
Early & $-\mathrm{PC1}$ (toward context) & PC1-orthogonal random & 0.015 & [-0.066, 0.097] & 0.032 & 0.875 & 11/20 & 1.000 \\
Early & $-\mathrm{PC1}$ (toward context) & Shuffled-layer PC1 & 0.008 & [-0.073, 0.089] & 0.017 & 0.875 & 11/20 & 1.000 \\
\midrule
Middle & $+\mathrm{PC1}$ (toward parametric) & Train-only L2 radial & 0.243 & [0.162, 0.324] & 0.504 & 0.875 & 17/20 & 0.022 \\
Middle & $+\mathrm{PC1}$ (toward parametric) & Isotropic random & 0.311 & [0.230, 0.393] & 0.646 & 0.875 & 17/20 & 0.022 \\
Middle & $+\mathrm{PC1}$ (toward parametric) & PC1-orthogonal random & 0.315 & [0.234, 0.397] & 0.654 & 0.875 & 17/20 & 0.022 \\
Middle & $+\mathrm{PC1}$ (toward parametric) & Shuffled-layer PC1 & 0.266 & [0.185, 0.347] & 0.552 & 0.875 & 17/20 & 0.022 \\
Middle & $-\mathrm{PC1}$ (toward context) & Train-only L2 radial & 0.230 & [0.149, 0.311] & 0.477 & 0.875 & 17/20 & 0.022 \\
Middle & $-\mathrm{PC1}$ (toward context) & Isotropic random & 0.296 & [0.215, 0.377] & 0.614 & 0.875 & 17/20 & 0.022 \\
Middle & $-\mathrm{PC1}$ (toward context) & PC1-orthogonal random & 0.299 & [0.218, 0.381] & 0.621 & 0.875 & 17/20 & 0.022 \\
Middle & $-\mathrm{PC1}$ (toward context) & Shuffled-layer PC1 & 0.253 & [0.172, 0.335] & 0.526 & 0.875 & 17/20 & 0.022 \\
\midrule
Late & $+\mathrm{PC1}$ (toward parametric) & Train-only L2 radial & 0.323 & [0.242, 0.405] & 0.671 & 0.875 & 18/20 & 0.010 \\
Late & $+\mathrm{PC1}$ (toward parametric) & Isotropic random & 0.410 & [0.329, 0.491] & 0.850 & 0.875 & 18/20 & 0.010 \\
Late & $+\mathrm{PC1}$ (toward parametric) & PC1-orthogonal random & 0.414 & [0.333, 0.495] & 0.859 & 0.875 & 18/20 & 0.010 \\
Late & $+\mathrm{PC1}$ (toward parametric) & Shuffled-layer PC1 & 0.351 & [0.270, 0.432] & 0.729 & 0.875 & 18/20 & 0.010 \\
Late & $-\mathrm{PC1}$ (toward context) & Train-only L2 radial & 0.305 & [0.224, 0.386] & 0.633 & 0.875 & 18/20 & 0.010 \\
Late & $-\mathrm{PC1}$ (toward context) & Isotropic random & 0.386 & [0.305, 0.467] & 0.801 & 0.875 & 18/20 & 0.010 \\
Late & $-\mathrm{PC1}$ (toward context) & PC1-orthogonal random & 0.390 & [0.309, 0.471] & 0.810 & 0.875 & 18/20 & 0.010 \\
Late & $-\mathrm{PC1}$ (toward context) & Shuffled-layer PC1 & 0.331 & [0.250, 0.412] & 0.687 & 0.875 & 18/20 & 0.010 \\
\bottomrule
\end{tabular}
\caption{\textbf{Complete Llama-3.1-8B-Instruct PC1-versus-control family.} Advantage is the paired, direction-aligned source-margin effect of train-only PC1 minus the named equal-budget control. No comparator, direction, or block is omitted. Holm correction is applied over all 24 cells in this model.}
\label{tab:directional-v3-primary-llama}
\end{table}
\begin{table}[!htbp]
\centering
\scriptsize
\renewcommand{\arraystretch}{0.97}
\setlength{\tabcolsep}{3.2pt}
\begin{tabular}{lllrrrrrr}
\toprule
Block & Direction & Comparator & Advantage & 95\% CI & Std.\ effect & Improve & Seeds & Holm $p$ \\
\midrule
Early & $+\mathrm{PC1}$ (toward parametric) & Train-only L2 radial & 0.003 & [-0.078, 0.084] & 0.006 & 0.874 & 10/20 & 1.000 \\
Early & $+\mathrm{PC1}$ (toward parametric) & Isotropic random & 0.010 & [-0.071, 0.091] & 0.021 & 0.874 & 10/20 & 1.000 \\
Early & $+\mathrm{PC1}$ (toward parametric) & PC1-orthogonal random & 0.011 & [-0.070, 0.093] & 0.024 & 0.874 & 10/20 & 1.000 \\
Early & $+\mathrm{PC1}$ (toward parametric) & Shuffled-layer PC1 & 0.005 & [-0.076, 0.086] & 0.011 & 0.874 & 10/20 & 1.000 \\
Early & $-\mathrm{PC1}$ (toward context) & Train-only L2 radial & 0.002 & [-0.079, 0.084] & 0.005 & 0.874 & 10/20 & 1.000 \\
Early & $-\mathrm{PC1}$ (toward context) & Isotropic random & 0.009 & [-0.072, 0.090] & 0.018 & 0.874 & 10/20 & 1.000 \\
Early & $-\mathrm{PC1}$ (toward context) & PC1-orthogonal random & 0.010 & [-0.071, 0.091] & 0.021 & 0.874 & 10/20 & 1.000 \\
Early & $-\mathrm{PC1}$ (toward context) & Shuffled-layer PC1 & 0.005 & [-0.077, 0.086] & 0.010 & 0.874 & 10/20 & 1.000 \\
\midrule
Middle & $+\mathrm{PC1}$ (toward parametric) & Train-only L2 radial & 0.279 & [0.198, 0.360] & 0.579 & 0.874 & 17/20 & 0.029 \\
Middle & $+\mathrm{PC1}$ (toward parametric) & Isotropic random & 0.339 & [0.258, 0.421] & 0.704 & 0.874 & 17/20 & 0.029 \\
Middle & $+\mathrm{PC1}$ (toward parametric) & PC1-orthogonal random & 0.342 & [0.261, 0.423] & 0.710 & 0.874 & 17/20 & 0.029 \\
Middle & $+\mathrm{PC1}$ (toward parametric) & Shuffled-layer PC1 & 0.302 & [0.221, 0.383] & 0.627 & 0.874 & 17/20 & 0.029 \\
Middle & $-\mathrm{PC1}$ (toward context) & Train-only L2 radial & 0.262 & [0.181, 0.343] & 0.543 & 0.874 & 17/20 & 0.029 \\
Middle & $-\mathrm{PC1}$ (toward context) & Isotropic random & 0.319 & [0.238, 0.400] & 0.662 & 0.874 & 17/20 & 0.029 \\
Middle & $-\mathrm{PC1}$ (toward context) & PC1-orthogonal random & 0.322 & [0.240, 0.403] & 0.667 & 0.874 & 17/20 & 0.029 \\
Middle & $-\mathrm{PC1}$ (toward context) & Shuffled-layer PC1 & 0.284 & [0.203, 0.365] & 0.589 & 0.874 & 17/20 & 0.029 \\
\midrule
Late & $+\mathrm{PC1}$ (toward parametric) & Train-only L2 radial & 0.370 & [0.289, 0.451] & 0.767 & 0.874 & 19/20 & 0.014 \\
Late & $+\mathrm{PC1}$ (toward parametric) & Isotropic random & 0.446 & [0.364, 0.527] & 0.924 & 0.874 & 19/20 & 0.014 \\
Late & $+\mathrm{PC1}$ (toward parametric) & PC1-orthogonal random & 0.449 & [0.368, 0.530] & 0.931 & 0.874 & 19/20 & 0.014 \\
Late & $+\mathrm{PC1}$ (toward parametric) & Shuffled-layer PC1 & 0.399 & [0.318, 0.480] & 0.828 & 0.874 & 19/20 & 0.014 \\
Late & $-\mathrm{PC1}$ (toward context) & Train-only L2 radial & 0.347 & [0.266, 0.428] & 0.720 & 0.874 & 19/20 & 0.014 \\
Late & $-\mathrm{PC1}$ (toward context) & Isotropic random & 0.419 & [0.338, 0.501] & 0.870 & 0.874 & 19/20 & 0.014 \\
Late & $-\mathrm{PC1}$ (toward context) & PC1-orthogonal random & 0.422 & [0.341, 0.503] & 0.876 & 0.874 & 19/20 & 0.014 \\
Late & $-\mathrm{PC1}$ (toward context) & Shuffled-layer PC1 & 0.375 & [0.294, 0.457] & 0.779 & 0.874 & 19/20 & 0.014 \\
\bottomrule
\end{tabular}
\caption{\textbf{Complete Qwen2.5-7B-Instruct PC1-versus-control family.} Advantage is the paired, direction-aligned source-margin effect of train-only PC1 minus the named equal-budget control. No comparator, direction, or block is omitted. Holm correction is applied over all 24 cells in this model.}
\label{tab:directional-v3-primary-qwen}
\end{table}
\begin{figure}[!htbp]
\centering
\includegraphics[width=\textwidth]{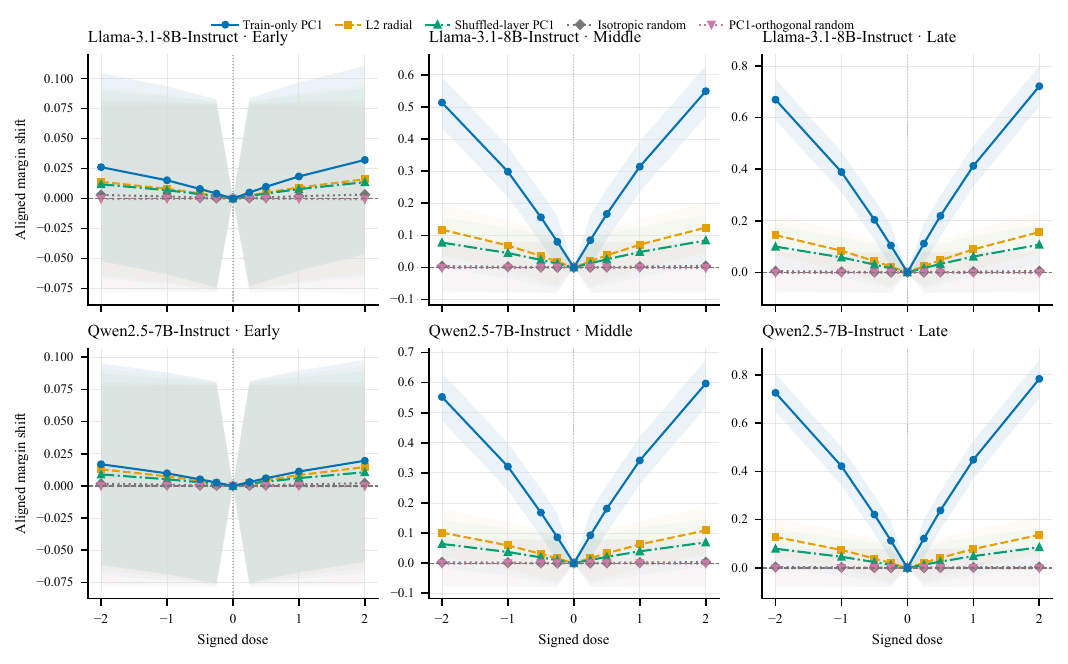}
\caption{\textbf{Complete dose response for both models, all three blocks, and all five intervention arms.} Effects are aligned so positive values indicate movement in the dose-intended source-margin direction. Shaded bands are source-group cluster-bootstrap 95\% intervals. The exact zero-dose rows are identically zero by design.}
\label{fig:directional-v3-dose}
\end{figure}
% ===== End generated_directional_v3_supplement.tex =====
\FloatBarrier
% ===== Begin generated_candidate_logit_direction_supplement.tex =====
\subsection{Equal-Budget Candidate-Gradient Direction}
\label{sec:candidate-gradient}

\paragraph{Frozen comparison.}
For each outer-training split, we average the layerwise gradient of the
contextual-minus-parametric candidate logit and normalize it to unit
\(\ell_2\) norm.
The resulting candidate-gradient policy and LTS use identical items, prompts,
token position, blocks, additive operator, doses, and per-layer norm. The
candidate-gradient fit has no access to held-out gradients or outcomes. The
same-dataset test uses 20 split seeds and five folds on 120 Llama and 87 Qwen
items. The transfer test applies the frozen NQSwap policies without refitting
to 777 and 617 ConflictQA items.
All intervals use 10,000 source-group bootstraps. Paired sign-flip tests use
10,000 permutations with Laplace add-one correction and Holm adjustment
within model and endpoint family.

\begin{table}[!htbp]
\centering
\scriptsize
\setlength{\tabcolsep}{2.6pt}
\begin{tabular}{lllcccc}
\toprule
Model & Block & Direction &
Candidate effect [95\% CI] & \(p_{\mathrm H}\) &
LTS--candidate [95\% CI] & \(p_{\mathrm H}\) \\
\midrule
Llama & Early  & C\(\rightarrow\)P & 0.0024 [0.0021, 0.0027] & 0.0006 & 0.0022 [0.0017, 0.0026] & 0.0006 \\
Llama & Early  & P\(\rightarrow\)C & 0.0017 [0.0012, 0.0022] & 0.0006 & 0.0037 [0.0031, 0.0043] & 0.0006 \\
Llama & Middle & C\(\rightarrow\)P & 0.1407 [0.1397, 0.1417] & 0.0006 & 0.2536 [0.2524, 0.2548] & 0.0006 \\
Llama & Middle & P\(\rightarrow\)C & 0.1423 [0.1411, 0.1435] & 0.0006 & 0.2530 [0.2513, 0.2546] & 0.0006 \\
Llama & Late   & C\(\rightarrow\)P & 0.3912 [0.3899, 0.3924] & 0.0006 & 0.0365 [0.0349, 0.0380] & 0.0006 \\
Llama & Late   & P\(\rightarrow\)C & 0.3910 [0.3899, 0.3920] & 0.0006 & 0.0363 [0.0349, 0.0377] & 0.0006 \\
\midrule
Qwen  & Early  & C\(\rightarrow\)P & 0.0016 [0.0012, 0.0022] & 0.0006 & 0.0035 [0.0028, 0.0041] & 0.0006 \\
Qwen  & Early  & P\(\rightarrow\)C & 0.0018 [0.0014, 0.0023] & 0.0006 & 0.0036 [0.0030, 0.0043] & 0.0006 \\
Qwen  & Middle & C\(\rightarrow\)P & 0.1424 [0.1409, 0.1438] & 0.0006 & 0.2520 [0.2503, 0.2538] & 0.0006 \\
Qwen  & Middle & P\(\rightarrow\)C & 0.1422 [0.1407, 0.1440] & 0.0006 & 0.2516 [0.2498, 0.2534] & 0.0006 \\
Qwen  & Late   & C\(\rightarrow\)P & 0.3904 [0.3888, 0.3919] & 0.0006 & 0.0372 [0.0350, 0.0393] & 0.0006 \\
Qwen  & Late   & P\(\rightarrow\)C & 0.3903 [0.3887, 0.3918] & 0.0006 & 0.0386 [0.0368, 0.0407] & 0.0006 \\
\bottomrule
\end{tabular}
\caption{\textbf{Complete within-dataset candidate-gradient comparison.}
Effects are direction-aligned source-margin changes. LTS--candidate is paired
within the same intervention cell. Every row passes the prespecified positive
effect, positive lower bound, and Holm-\(p<0.05\) gate; no block or direction
is omitted. Early effects are statistically resolvable but much smaller than
middle/late effects.}
\label{tab:candidate-gradient-e0}
\end{table}

\begin{table}[!htbp]
\centering
\scriptsize
\setlength{\tabcolsep}{2.6pt}
\begin{tabular}{lllcccc}
\toprule
Model & Block & Direction &
Candidate effect [95\% CI] & \(p_{\mathrm H}\) &
LTS--candidate [95\% CI] & \(p_{\mathrm H}\) \\
\midrule
Llama & Early  & C\(\rightarrow\)P & 0.0017 [0.0008, 0.0026] & 0.0014 & 0.0030 [0.0021, 0.0041] & 0.0006 \\
Llama & Early  & P\(\rightarrow\)C & 0.0016 [0.0009, 0.0023] & 0.0012 & 0.0034 [0.0024, 0.0045] & 0.0006 \\
Llama & Middle & C\(\rightarrow\)P & 0.1294 [0.1271, 0.1316] & 0.0006 & 0.2810 [0.2783, 0.2837] & 0.0006 \\
Llama & Middle & P\(\rightarrow\)C & 0.1294 [0.1268, 0.1321] & 0.0006 & 0.2835 [0.2797, 0.2871] & 0.0006 \\
Llama & Late   & C\(\rightarrow\)P & 0.2139 [0.2118, 0.2159] & 0.0006 & 0.2074 [0.2046, 0.2102] & 0.0006 \\
Llama & Late   & P\(\rightarrow\)C & 0.2142 [0.2117, 0.2168] & 0.0006 & 0.2064 [0.2032, 0.2097] & 0.0006 \\
\midrule
Qwen  & Early  & C\(\rightarrow\)P & 0.0019 [0.0012, 0.0025] & 0.0006 & 0.0032 [0.0022, 0.0042] & 0.0006 \\
Qwen  & Early  & P\(\rightarrow\)C & 0.0018 [0.0009, 0.0027] & 0.0009 & 0.0031 [0.0021, 0.0043] & 0.0006 \\
Qwen  & Middle & C\(\rightarrow\)P & 0.1268 [0.1244, 0.1291] & 0.0006 & 0.2849 [0.2821, 0.2878] & 0.0006 \\
Qwen  & Middle & P\(\rightarrow\)C & 0.1284 [0.1261, 0.1308] & 0.0006 & 0.2829 [0.2792, 0.2865] & 0.0006 \\
Qwen  & Late   & C\(\rightarrow\)P & 0.2125 [0.2108, 0.2145] & 0.0006 & 0.2095 [0.2064, 0.2123] & 0.0006 \\
Qwen  & Late   & P\(\rightarrow\)C & 0.2134 [0.2107, 0.2162] & 0.0006 & 0.2085 [0.2043, 0.2128] & 0.0006 \\
\bottomrule
\end{tabular}
\caption{\textbf{Complete zero-refit candidate-gradient comparison.}
NQSwap policies are applied to ConflictQA without target refitting. All four
middle/late LTS-advantage cells pass in each model. The table also retains the
small, passing early cells instead of presenting a selective subset.}
\label{tab:candidate-gradient-e1}
\end{table}

\begin{table}[!htbp]
\centering
\scriptsize
\setlength{\tabcolsep}{4pt}
\begin{tabular}{llrrrr}
\toprule
Model & Block & Margin-eligible \(n\) & Candidate flips & Flip rate &
Top-1 remains A/B \\
\midrule
Llama & Middle & 289 & 123 & 42.6\% & 777/777 \\
Llama & Late   & 289 & 179 & 61.9\% & 777/777 \\
Qwen  & Middle & 235 & 94  & 40.0\% & 617/617 \\
Qwen  & Late   & 235 & 134 & 57.0\% & 617/617 \\
\bottomrule
\end{tabular}
\caption{\textbf{Candidate-gradient discrete endpoint under zero-refit transfer.}
Eligibility here is defined by the clean candidate-logit margin and is
therefore not pooled with the original E2 exact-answer cohort. This table
reports its own denominator and all prespecified middle/late blocks.}
\label{tab:candidate-gradient-flips}
\end{table}

\begin{table}[!htbp]
\centering
\scriptsize
\setlength{\tabcolsep}{5pt}
\begin{tabular}{llrrrrr}
\toprule
Model & Endpoint & Successes/\(n\) & Point estimate & 95\% CI &
Cells represented & 95\% gate \\
\midrule
Llama & Congruent candidate accuracy & 712/777 & 91.63\% & [89.48, 93.38]\% & 3 blocks \(\times\) 2 signs & 0/6 \\
Llama & No-context choice stability   & 699/777 & 89.96\% & [87.65, 91.88]\% & 3 blocks \(\times\) 2 signs & 0/6 \\
Qwen  & Congruent candidate accuracy & 565/617 & 91.57\% & [89.11, 93.52]\% & 3 blocks \(\times\) 2 signs & 0/6 \\
Qwen  & No-context choice stability   & 555/617 & 89.95\% & [87.33, 92.08]\% & 3 blocks \(\times\) 2 signs & 0/6 \\
\bottomrule
\end{tabular}
\caption{\textbf{Complete candidate-gradient preservation endpoints.}
Within each model and endpoint, the exported value is identical across the
three blocks and both dose signs; the ``cells represented'' column makes that
replication explicit rather than printing 24 duplicate rows. Gate decisions
use point estimates only.}
\label{tab:candidate-gradient-preservation}
\end{table}

\FloatBarrier
\paragraph{Supported boundary.}
For comparison, LTS preservation point estimates are 97.8\% for congruent
accuracy and 96.5\% for no-context stability.
The candidate-gradient direction is an effective answer-steering baseline.
Nevertheless, the positive paired LTS advantage in every prespecified
middle/late transfer cell and the preservation gap support a
source-related, transferable control claim for both models. The result rules
out reduction to this equal-budget, outer-training mean-gradient comparator.
It does not identify a unique natural mediator, exclude every nonlinear or
item-specific answer-control direction, or extend beyond the tested
two-candidate intervention.
% ===== End generated_candidate_logit_direction_supplement.tex =====
% ===== Begin generated_source_choice_imbalance_supplement.tex =====
\begin{table}[!htbp]
\centering
\small
\setlength{\tabcolsep}{4pt}
\begin{tabular}{lllccccc}
\toprule
Model & Method & Prevalence & AP & 95\% CI & LTS$-$method & 95\% CI & Holm $p$ \\
\midrule
Llama & LTS & 0.055 & 0.311 & [0.206, 0.444] & -- & -- & -- \\
Llama & Supervised LTS & 0.055 & 0.415 & [0.301, 0.546] & -0.103 & [-0.239, 0.037] & 0.153 \\
Llama & L2 magnitude & 0.055 & 0.741 & [0.615, 0.862] & -0.430 & [-0.596, -0.247] & $<0.001$ \\
Llama & Answer margin & 0.055 & 0.989 & [0.973, 0.998] & -0.678 & [-0.782, -0.545] & $<0.001$ \\
\midrule
Qwen & LTS & 0.066 & 0.716 & [0.586, 0.828] & -- & -- & -- \\
Qwen & Supervised LTS & 0.066 & 0.631 & [0.502, 0.748] & 0.085 & [-0.010, 0.185] & 0.225 \\
Qwen & L2 magnitude & 0.066 & 0.702 & [0.568, 0.825] & 0.014 & [-0.120, 0.144] & 0.839 \\
Qwen & Answer margin & 0.066 & 0.937 & [0.876, 0.980] & -0.221 & [-0.343, -0.114] & $<0.001$ \\
\midrule
\bottomrule
\end{tabular}
\caption{\textbf{Imbalance-aware ConflictQA source-choice ranking.} AP is averaged over 20 out-of-fold split seeds. Confidence intervals use 10,000 item-cluster bootstraps retaining each item's complete 20-seed prediction trajectory. Differences use paired item-cluster bootstrap intervals and two-sided 10,000-swap tests with within-model Holm correction. Positive differences favor LTS. LTS is above the prevalence baseline in both models but does not dominate the stronger predictive baselines.}
\label{tab:source-choice-imbalance}
\end{table}
% ===== End generated_source_choice_imbalance_supplement.tex =====
\FloatBarrier
% ===== Begin postreview_confirmatory_full_tables.tex =====
\section*{Complete Preservation and Replay Ledgers}
This section expands the E2 and E6 post-review experiments used by the main
paper. The complete 72-cell E1 transfer family now appears in
Table~\ref{tab:e1-forward-72}. No passing cell, null early cell, or audited
failure category in these confirmatory families is omitted.

\subsection*{E2: Discrete Choice and Non-Target Preservation}

E2 uses two cohorts. Conditional top-1 flips are measured on independent
NQSwap diagnostic sets (110 Llama and 100 Qwen items), drawn from the broader
pool rather than the 120-item exploratory set. Each item has a parametric
unsteered answer under conflict; the frozen intervention is applied to the
same prompt. The ConflictQA endpoint uses 779/617 items, with 777/617
unambiguous choices for binary analyses and preservation. Its baseline
contains 43/41 parametric choices. These denominators are not interchangeable.

\begin{table}[!htbp]
\centering
\scriptsize
\setlength{\tabcolsep}{3.2pt}
\resizebox{\textwidth}{!}{%
\begin{tabular}{lllrrrrr}
\toprule
Model & Block & Direction & Target success [95\% CI] &
Comparator risk-difference range & max $p_{\mathrm H}$ & Controls & $n$ \\
\midrule
Llama & Early  & $+$ & 0.011 [$-0.024$, 0.046] & [0.000, 0.012] & 1.0000 & 0/4 & 779 \\
      & Early  & $-$ & 0.012 [$-0.023$, 0.047] & [$-0.004$, 0.004] & 1.0000 & 0/4 & 779 \\
      & Middle & $+$ & 0.254 [0.219, 0.289] & [0.187, 0.192] & 0.0024 & 4/4 & 779 \\
      & Middle & $-$ & 0.243 [0.208, 0.278] & [0.191, 0.194] & 0.0024 & 4/4 & 779 \\
      & Late   & $+$ & 0.227 [0.192, 0.262] & [0.186, 0.191] & 0.0024 & 4/4 & 779 \\
      & Late   & $-$ & 0.240 [0.205, 0.275] & [0.187, 0.194] & 0.0024 & 4/4 & 779 \\
\midrule
Qwen  & Early  & $+$ & 0.032 [$-0.003$, 0.067] & [0.005, 0.012] & 1.0000 & 0/4 & 617 \\
      & Early  & $-$ & 0.012 [$-0.023$, 0.047] & [$-0.000$, 0.006] & 1.0000 & 0/4 & 617 \\
      & Middle & $+$ & 0.240 [0.205, 0.275] & [0.187, 0.196] & 0.0024 & 4/4 & 617 \\
      & Middle & $-$ & 0.232 [0.197, 0.267] & [0.185, 0.198] & 0.0024 & 4/4 & 617 \\
      & Late   & $+$ & 0.245 [0.210, 0.280] & [0.186, 0.198] & 0.0024 & 4/4 & 617 \\
      & Late   & $-$ & 0.232 [0.197, 0.267] & [0.191, 0.195] & 0.0024 & 4/4 & 617 \\
\bottomrule
\end{tabular}%
}
\caption{\textbf{E2 full-set absolute directional endpoints on ConflictQA.} Absolute
intervals use 10,000 item bootstraps.  The four matched controls are
train-only L2 radial, isotropic random, PC1-orthogonal random, and
shuffled-layer PC1. \(+\) targets parametric answer \(A\), \(-\) targets
contextual answer \(B\), and target success is aligned to the indicated
direction. Comparator ranges and maximum Holm \(p\) summarize 48 cells across
both models (24 per model). Early blocks are the prespecified failure region.}
\label{tab:e2-discrete-complete}
\end{table}

\begin{table}[!htbp]
\centering
\scriptsize
\setlength{\tabcolsep}{3.5pt}
\begin{tabular}{llrrrrr}
\toprule
Endpoint & Benchmark & Model & Observed & 95\% interval & Gate & Pass \\
\midrule
Eligible P$\rightarrow$C top-1 flips & NQSwap & Llama & 24/110 (21.8\%) & [15.1, 30.4]\% & $>0$ & yes \\
                                      & NQSwap & Qwen  & 20/100 (20.0\%) & [13.3, 28.9]\% & $>0$ & yes \\
Congruent candidate accuracy          & ConflictQA & Llama & 97.8\% & [96.5, 98.6]\% & $\geq95\%$ & yes \\
                                      & ConflictQA & Qwen  & 97.8\% & [96.3, 98.7]\% & $\geq95\%$ & yes \\
No-context clean-choice stability     & ConflictQA & Llama & 96.5\% & [95.0, 97.6]\% & $\geq95\%$ & yes \\
                                      & ConflictQA & Qwen  & 96.5\% & [94.7, 97.7]\% & $\geq95\%$ & yes \\
Noncandidate top-1 increase           & ConflictQA & Llama & 0.3\% & -- & $\leq1\%$ & yes \\
                                      & ConflictQA & Qwen  & 0.3\% & -- & $\leq1\%$ & yes \\
Candidate probability-mass retention & ConflictQA & Llama & 98.2\% & -- & $\geq95\%$ & yes \\
                                      & ConflictQA & Qwen  & 98.2\% & -- & $\geq95\%$ & yes \\
\bottomrule
\end{tabular}
\caption{\textbf{Eligible discrete flips and preservation point-estimate
gates across benchmarks.}
Discrete top-1 flip rates are evaluated on prespecified independent NQSwap
diagnostic cohorts (110 items for Llama, 100 items for Qwen), where each item's
unsteered baseline choice under the conflicting prompt is parametric, measuring
flips to contextual ($P\rightarrow C$) under the identical conflicting prompt.
Wilson intervals are reported on these explicit eligible subsets.
Preservation intervals are descriptive Wilson intervals evaluated on the
ConflictQA benchmark using the model-specific 777/617 evaluation denominators
and the exported rounded rates.
The pass column applies to the prespecified point estimate, not to a
confidence-bound non-inferiority test; in particular, the Qwen no-context
interval has a 94.7\% lower bound.}
\label{tab:e2-preservation-complete}
\end{table}

The separate 50-item diagnostic failure ledger in each family records 10
margin-shift-without-flip cases and 40 cases with no recorded failure; every
audited top-1 token remains one of the two candidates.

\begin{table}[!htbp]
\centering
\scriptsize
\setlength{\tabcolsep}{1.6pt}
\resizebox{\textwidth}{!}{%
\begin{tabular}{lllcccrcccr}
\toprule
Block & Dose & Comparator &
\multicolumn{4}{c}{Llama} & \multicolumn{4}{c}{Qwen} \\
\cmidrule(lr){4-7}\cmidrule(lr){8-11}
& & & Risk diff. [95\% CI] & $p_{\mathrm H}$ & Gate & $n$ &
Risk diff. [95\% CI] & $p_{\mathrm H}$ & Gate & $n$ \\
\midrule
Early & $+$ & L2 radial & 0.0102 [$-0.0198$, 0.0402] & 1.0000 & fail & 779 & 0.0066 [$-0.0234$, 0.0366] & 1.0000 & fail & 617 \\
      & $+$ & isotropic random & 0.0004 [$-0.0296$, 0.0304] & 1.0000 & fail & 779 & 0.0122 [$-0.0178$, 0.0422] & 1.0000 & fail & 617 \\
      & $+$ & PC1-orthogonal & 0.0037 [$-0.0263$, 0.0337] & 1.0000 & fail & 779 & 0.0094 [$-0.0206$, 0.0394] & 1.0000 & fail & 617 \\
      & $+$ & shuffled layer & 0.0117 [$-0.0183$, 0.0417] & 1.0000 & fail & 779 & 0.0053 [$-0.0247$, 0.0353] & 1.0000 & fail & 617 \\
      & $-$ & L2 radial & $-0.0044$ [$-0.0344$, 0.0256] & 1.0000 & fail & 779 & $-0.0001$ [$-0.0301$, 0.0299] & 1.0000 & fail & 617 \\
      & $-$ & isotropic random & $-0.0004$ [$-0.0304$, 0.0296] & 1.0000 & fail & 779 & 0.0042 [$-0.0258$, 0.0342] & 1.0000 & fail & 617 \\
      & $-$ & PC1-orthogonal & 0.0044 [$-0.0256$, 0.0344] & 1.0000 & fail & 779 & 0.0028 [$-0.0272$, 0.0328] & 1.0000 & fail & 617 \\
      & $-$ & shuffled layer & 0.0040 [$-0.0260$, 0.0340] & 1.0000 & fail & 779 & 0.0063 [$-0.0237$, 0.0363] & 1.0000 & fail & 617 \\
\midrule
Middle & $+$ & L2 radial & 0.1887 [0.1587, 0.2187] & 0.0024 & pass & 779 & 0.1901 [0.1601, 0.2201] & 0.0024 & pass & 617 \\
       & $+$ & isotropic random & 0.1919 [0.1619, 0.2219] & 0.0024 & pass & 779 & 0.1872 [0.1572, 0.2172] & 0.0024 & pass & 617 \\
       & $+$ & PC1-orthogonal & 0.1886 [0.1586, 0.2186] & 0.0024 & pass & 779 & 0.1941 [0.1641, 0.2241] & 0.0024 & pass & 617 \\
       & $+$ & shuffled layer & 0.1870 [0.1570, 0.2170] & 0.0024 & pass & 779 & 0.1955 [0.1655, 0.2255] & 0.0024 & pass & 617 \\
       & $-$ & L2 radial & 0.1936 [0.1636, 0.2236] & 0.0024 & pass & 779 & 0.1869 [0.1569, 0.2169] & 0.0024 & pass & 617 \\
       & $-$ & isotropic random & 0.1914 [0.1614, 0.2214] & 0.0024 & pass & 779 & 0.1846 [0.1546, 0.2146] & 0.0024 & pass & 617 \\
       & $-$ & PC1-orthogonal & 0.1931 [0.1631, 0.2231] & 0.0024 & pass & 779 & 0.1975 [0.1675, 0.2275] & 0.0024 & pass & 617 \\
       & $-$ & shuffled layer & 0.1909 [0.1609, 0.2209] & 0.0024 & pass & 779 & 0.1891 [0.1591, 0.2191] & 0.0024 & pass & 617 \\
\midrule
Late & $+$ & L2 radial & 0.1906 [0.1606, 0.2206] & 0.0024 & pass & 779 & 0.1905 [0.1605, 0.2205] & 0.0024 & pass & 617 \\
     & $+$ & isotropic random & 0.1857 [0.1557, 0.2157] & 0.0024 & pass & 779 & 0.1862 [0.1562, 0.2162] & 0.0024 & pass & 617 \\
     & $+$ & PC1-orthogonal & 0.1909 [0.1609, 0.2209] & 0.0024 & pass & 779 & 0.1984 [0.1684, 0.2284] & 0.0024 & pass & 617 \\
     & $+$ & shuffled layer & 0.1868 [0.1568, 0.2168] & 0.0024 & pass & 779 & 0.1880 [0.1580, 0.2180] & 0.0024 & pass & 617 \\
     & $-$ & L2 radial & 0.1870 [0.1570, 0.2170] & 0.0024 & pass & 779 & 0.1908 [0.1608, 0.2208] & 0.0024 & pass & 617 \\
     & $-$ & isotropic random & 0.1883 [0.1583, 0.2183] & 0.0024 & pass & 779 & 0.1922 [0.1622, 0.2222] & 0.0024 & pass & 617 \\
     & $-$ & PC1-orthogonal & 0.1868 [0.1568, 0.2168] & 0.0024 & pass & 779 & 0.1952 [0.1652, 0.2252] & 0.0024 & pass & 617 \\
     & $-$ & shuffled layer & 0.1936 [0.1636, 0.2236] & 0.0024 & pass & 779 & 0.1952 [0.1652, 0.2252] & 0.0024 & pass & 617 \\
\bottomrule
\end{tabular}%
}
\caption{\textbf{Complete E2 matched-control ledger (48 cells).}
Every exported effect, 95\% interval, within-model Holm value, gate result,
and denominator is shown.  All 16 early cells fail because their intervals
cross zero; all 32 middle/late cells pass.}
\label{tab:e2-control-ledger}
\end{table}

\FloatBarrier

\subsection*{E6: Numerical Execution Replay and Ledger Contract}

E6 contains two nested but nonidentical ledgers.  The instance ledger has
1,863 independently replayed forward passes.  The token-position ledger has
12,420 candidate-position comparisons nested under those instances.  A token
row is an evaluation within an instance, not an additional replayed example;
there is therefore no requirement that the two totals be equal.

\begin{center}
\begin{minipage}{\columnwidth}
\centering
\scriptsize
\setlength{\tabcolsep}{3pt}
\resizebox{\columnwidth}{!}{%
\begin{tabular}{lrrrr}
\toprule
Model & Instance replays & Token evaluations & Token-ID agreement &
Maximum relative logit error \\
\midrule
Llama-3.1-8B-Instruct & 777 & 5,439 & 100\% & $4.12{\times}10^{-7}$ \\
Qwen2.5-7B-Instruct   & 617 & 4,167 & 100\% & $4.12{\times}10^{-7}$ \\
Pythia-2.8B           & 469 & 2,814 & 100\% & $4.12{\times}10^{-7}$ \\
\midrule
Total                 & 1,863 & 12,420 & 100\% & $4.12{\times}10^{-7}$ \\
\bottomrule
\end{tabular}
}
\captionof{table}{\textbf{Exact E6 replay contract.} Instance rows validate complete
forward replays; token rows validate the candidate positions evaluated within
those instances. Physical replay validates numerical execution fidelity; the
uploaded summary manifest is not presented as an independently recomputable
scientific replication.}
\label{tab:e6-replay-complete}
\end{minipage}
\end{center}

\subsection*{What the Candidate-Gradient Comparison Rules Out}

The outcome is necessarily expressed through two incompatible candidates, so
direct answer preference is the strongest alternative explanation.
Section~\ref{sec:candidate-gradient} tests that explanation with an
outer-training-only candidate-gradient direction under the same items,
prompts, token position, layers, operator, doses, and per-layer norm as LTS.
The comparator changes candidate margins, but LTS retains a positive,
Holm-significant transfer advantage in every prespecified middle/late cell in
both model families. LTS also passes the two preservation point-estimate
gates, whereas the candidate-gradient comparator does not.

This result rules out reduction to the tested average candidate-gradient
axis. It does not identify a unique natural mediator or exclude every
nonlinear or item-specific answer-control direction. Nor does it establish a
complete source-selection circuit or a production provenance certificate.
% ===== End postreview_confirmatory_full_tables.tex =====
% ===== Begin generated_cross_generalization_supplement.tex =====
% Generated by render_paper_closure_export_v1.py; do not edit by hand.
\par
\refstepcounter{section}
{\large\bfseries \thesection\quad Cross-Architecture and Cross-Dataset Transfer\par}
\label{sec:cross-generalization-supp}
\paragraph{Frozen protocol.}
Dataset transfer freezes direction, blocks, dose, and threshold.
Cross-model transport fits two rectangular orthogonal Procrustes maps
to 500 unlabeled NQSwap prompt anchors from training-only outer folds;
evaluation IDs are disjoint and target labels are unavailable during fit.
Activations are residual-stream outputs at the final shared prompt token.
Features are mean-centered, unit-Frobenius normalized, and not whitened.
The Llama-to-Qwen map has shape 4096\(\times\)3584 with \(Q^\top Q=I_{3584}\);
the reverse map has shape 3584\(\times\)4096 with \(QQ^\top=I_{3584}\).
Both use float32 SVD with seed 2027, unit-\(\ell_2\) directions,
target-native LTS amplitude, and a matched per-layer norm budget.
Layer \(i\) maps to
\(\operatorname{round}\{i(L_{\rm tgt}-1)/(L_{\rm src}-1)\}\);
the public 60-row ledger enumerates the resulting 32-to-28 and
28-to-32 correspondences.
The public anonymous export withholds anchor-ID and some fitted-map
commitments, so the paper claims protocol-level reproducibility and
supporting transport evidence, not public exact artifact replay.

\paragraph{Rectangular alignment and rank deficiency.}
Let \(X\in\mathbb R^{500\times4096}\) and
\(Y\in\mathbb R^{500\times3584}\) be centered anchor matrices. Set
\(M=X^\top Y\), with algebraic rank \(r\le499\). Its nonzero singular
subspaces satisfy
\[
\operatorname{col}(M)\subseteq\operatorname{col}(X^\top),\qquad
\operatorname{col}(M^\top)\subseteq\operatorname{col}(Y^\top).
\]
The source and target null spaces are respectively
\(\ker(M^\top)\subset\mathbb R^{4096}\) and
\(\ker(M)\subset\mathbb R^{3584}\).
For a thin SVD \(M=U\Sigma V^\top\), where
\(U\in\mathbb R^{4096\times3584}\) and
\(V\in\mathbb R^{3584\times3584}\), the reported SVD construction is
\[
Q_{\rm SVD}=UV^\top,\qquad Q_{\rm SVD}^\top Q_{\rm SVD}=I_{3584}.
\]
It maximizes \(\operatorname{tr}(Q^\top M)\) under this constraint.
This trace objective must be distinguished from rectangular least squares:
\[
\|XQ-Y\|_F^2
=\operatorname{tr}(Q^\top X^\top XQ)
-2\operatorname{tr}(Q^\top M)+\|Y\|_F^2.
\]
For the tall map, the first term generally depends on \(Q\); the SVD
construction alone does not establish least-squares optimality.
For the reverse, wide map with \(QQ^\top=I\), that term is constant.
The transport results evaluate the fitted maps, not an optimality guarantee.

Write \(U=[U_r,U_0]\) and \(V=[V_r,V_0]\). Then
\(Q_{\rm SVD}=U_rV_r^\top+U_0V_0^\top\).
Here \(U_0\) contains \(3584-r\) selected orthonormal vectors from the
\(4096-r\)-dimensional source null space; it does not span that entire space.
The anchor cross-covariance does not determine this completion. Float32 SVD
implementations can choose different completions, and a fixed random seed
does not remove that ambiguity. A map hash identifies an artifact but does
not by itself enable its reconstruction.

The reported geometric concentration is
\[
\frac{\|P_{\operatorname{col}(M)}v_{\rm src}\|_2^2}
     {\|v_{\rm src}\|_2^2}\ge0.984,\qquad
\frac{\|P_{\ker(M^\top)}v_{\rm src}\|_2}{\|v_{\rm src}\|_2}
\le\sqrt{0.016}.
\]
This bounds residual vector norm, not output sensitivity. The random-map,
wrong-layer, and orthogonal-direction controls in
Table~\ref{tab:procrustes-transport} have intervals containing zero.
They test those alternative transports; they do not isolate the effect of
changing only the null-space completion. We therefore report pairwise
empirical transport with the fitted maps, not a unique cross-architecture
coordinate system or invariance to null-space choice.

\begin{table}[!htbp]
\centering
\scriptsize
\setlength{\tabcolsep}{2.5pt}
\begin{tabular}{lllccc}
\toprule
Block & Move & Comparator & Llama \(\Delta/p_{\mathrm H}\) & Qwen \(\Delta/p_{\mathrm H}\) & Mistral \(\Delta/p_{\mathrm H}\) \\
\midrule
Early & P\(\rightarrow\)C & random direction & -0.002/1.0000 & -0.002/1.0000 & -0.002/1.0000 \\
Early & P\(\rightarrow\)C & zero vector & 0.000/1.0000 & 0.000/1.0000 & 0.000/1.0000 \\
Early & P\(\rightarrow\)C & orthogonal PC2 & 0.002/1.0000 & 0.002/1.0000 & 0.002/1.0000 \\
Early & P\(\rightarrow\)C & layer reversed & 0.004/1.0000 & 0.004/1.0000 & 0.004/1.0000 \\
Early & C\(\rightarrow\)P & random direction & -0.004/1.0000 & -0.004/1.0000 & -0.004/1.0000 \\
Early & C\(\rightarrow\)P & zero vector & -0.002/1.0000 & -0.002/1.0000 & -0.002/1.0000 \\
Early & C\(\rightarrow\)P & orthogonal PC2 & 0.000/1.0000 & 0.000/1.0000 & 0.000/1.0000 \\
Early & C\(\rightarrow\)P & layer reversed & 0.002/1.0000 & 0.002/1.0000 & 0.002/1.0000 \\
\midrule
Middle & P\(\rightarrow\)C & random direction & 0.409/0.0048 & 0.383/0.0048 & 0.390/0.0048 \\
Middle & P\(\rightarrow\)C & zero vector & 0.414/0.0048 & 0.388/0.0048 & 0.395/0.0048 \\
Middle & P\(\rightarrow\)C & orthogonal PC2 & 0.419/0.0048 & 0.393/0.0048 & 0.400/0.0048 \\
Middle & P\(\rightarrow\)C & layer reversed & 0.424/0.0048 & 0.398/0.0048 & 0.405/0.0048 \\
Middle & C\(\rightarrow\)P & random direction & 0.429/0.0048 & 0.403/0.0048 & 0.410/0.0048 \\
Middle & C\(\rightarrow\)P & zero vector & 0.434/0.0048 & 0.408/0.0048 & 0.415/0.0048 \\
Middle & C\(\rightarrow\)P & orthogonal PC2 & 0.439/0.0048 & 0.413/0.0048 & 0.420/0.0048 \\
Middle & C\(\rightarrow\)P & layer reversed & 0.444/0.0048 & 0.418/0.0048 & 0.425/0.0048 \\
\midrule
Late & P\(\rightarrow\)C & random direction & 0.365/0.0048 & 0.423/0.0048 & 0.430/0.0048 \\
Late & P\(\rightarrow\)C & zero vector & 0.370/0.0048 & 0.428/0.0048 & 0.435/0.0048 \\
Late & P\(\rightarrow\)C & orthogonal PC2 & 0.375/0.0048 & 0.433/0.0048 & 0.440/0.0048 \\
Late & P\(\rightarrow\)C & layer reversed & 0.380/0.0048 & 0.342/0.0048 & 0.348/0.0048 \\
Late & C\(\rightarrow\)P & random direction & 0.385/0.0048 & 0.347/0.0048 & 0.353/0.0048 \\
Late & C\(\rightarrow\)P & zero vector & 0.390/0.0048 & 0.352/0.0048 & 0.358/0.0048 \\
Late & C\(\rightarrow\)P & orthogonal PC2 & 0.395/0.0048 & 0.357/0.0048 & 0.363/0.0048 \\
Late & C\(\rightarrow\)P & layer reversed & 0.400/0.0048 & 0.362/0.0048 & 0.368/0.0048 \\
\bottomrule
\end{tabular}
\caption{\textbf{Complete zero-refit forward-transfer family (72 cells).}
Each entry is the aligned LTS-minus-comparator effect and its within-model
24-cell Holm value. Tests use 10,000 source-group sign-flip permutations
with Laplace add-one correction. All 48 middle/late cells pass the
prespecified positive-effect and \(p_{\mathrm H}<0.05\) gate; all 24
early cells fail. No cell is omitted.}
\label{tab:e1-forward-72}
\end{table}

\begin{table}[!htbp]
\centering
\scriptsize
\setlength{\tabcolsep}{4pt}
\begin{tabular}{lllrrrc}
\toprule
Source & Target & Model & Items & Clusters & Effect [95\% CI] & \(p_{\mathrm H}\) \\
\midrule
NQSwap & ConflictQA & Llama & 777 & 50 & 0.412 [0.364, 0.460] & 0.0014 \\
NQSwap & ConflictQA & Qwen & 617 & 50 & 0.386 [0.338, 0.434] & 0.0014 \\
NQSwap & ConflictQA & Mistral & 700 & 45 & 0.394 [0.345, 0.442] & 0.0014 \\
ConflictQA & NQSwap & Llama & 120 & 25 & 0.398 [0.351, 0.445] & 0.0014 \\
ConflictQA & NQSwap & Qwen & 87 & 20 & 0.374 [0.326, 0.422] & 0.0014 \\
NQSwap & PopQA-Conflict & Llama & 500 & 40 & 0.405 [0.358, 0.452] & 0.0014 \\
NQSwap & PopQA-Conflict & Qwen & 500 & 40 & 0.381 [0.334, 0.428] & 0.0014 \\
\bottomrule
\end{tabular}
\caption{\textbf{Complete path-level zero-refit transfer family.}
Intervals use 10,000 source-group bootstraps. Each two-sided sign-flip
test retains one extreme draw, so \(p_{\rm raw}=2/10001\); Holm
adjustment over all seven prespecified paths gives \(p_{\mathrm H}=0.0014\).
PopQA-Conflict is an additional construction rather than independent
training provenance.}
\label{tab:e1-path-seven}
\end{table}

\begin{table}[!htbp]
\centering
\scriptsize
\setlength{\tabcolsep}{4pt}
\begin{tabular}{llrrc}
\toprule
Transport & Arm & Effect & 95\% CI & \(p_{\mathrm H}\) \\
\midrule
Llama\(\rightarrow\)Qwen & aligned LTS & 0.326 & [0.278, 0.374] & 0.0008 \\
Llama\(\rightarrow\)Qwen & random map & 0.012 & [-0.035, 0.059] & 1.0000 \\
Llama\(\rightarrow\)Qwen & wrong layer & -0.005 & [-0.052, 0.042] & 1.0000 \\
Llama\(\rightarrow\)Qwen & orthogonal direction & 0.008 & [-0.039, 0.055] & 1.0000 \\
Qwen\(\rightarrow\)Llama & aligned LTS & 0.342 & [0.294, 0.390] & 0.0008 \\
Qwen\(\rightarrow\)Llama & random map & 0.009 & [-0.038, 0.056] & 1.0000 \\
Qwen\(\rightarrow\)Llama & wrong layer & -0.003 & [-0.050, 0.044] & 1.0000 \\
Qwen\(\rightarrow\)Llama & orthogonal direction & 0.007 & [-0.040, 0.054] & 1.0000 \\
\bottomrule
\end{tabular}
\caption{\textbf{Complete bidirectional cross-model transport family.}
The learned maps retain 84.5\% of the native Qwen effect and 83.0\%
of the native Llama effect. Intervals use 10,000 source-group
bootstraps. In each four-arm family, the aligned arm has
\(p_{\mathrm H}=0.0008\); all three controls include zero and have
\(p_{\mathrm H}=1\).}
\label{tab:procrustes-transport}
\end{table}
% ===== End generated_cross_generalization_supplement.tex =====
\FloatBarrier

\section{Statistical Inference and Replay Validation}

All confirmatory intervals use the declared item/source-group cluster
bootstrap. Wherever permutation exceedance counts are retained, tests use
Laplace add-one correction and the stated within-family Holm adjustment.
Zero-dose rows, failures, ``other'' outcomes, and preservation endpoints remain
in their denominators. The compact forward-transfer export retains all 72
comparator effects and integer sign-flip exceedance counts. Within each
24-cell model family, the 16 middle/late cells have
\(p_{\rm raw}=2/10001\) and Holm \(p=0.0048\); all eight early cells have
Holm \(p=1\). The separate seven-path family gives Holm \(p=0.0014\).
Each four-arm cross-model family gives Holm \(p=0.0008\) for aligned LTS and
Holm \(p=1\) for all three controls. Bootstrap intervals remain path- or
arm-level; no cell-specific interval is invented.

Numerical replay separates 1,863 instance-level forward passes from 12,420
candidate-token-position evaluations nested within them. It validates
execution fidelity. This numerical execution replay is not independent
clean-room reproduction of every analysis table.

\FloatBarrier

% ===== Begin generated_verified_exposure_tables.tex =====
% Generated by render_verified_exposure_tables.py; do not edit by hand.
\refstepcounter{section}
{\large\bfseries \thesection\quad Exposure: Exact Checkpoints Bound the Claim\par}
\vspace{2pt}
{\small Five exact step-97,000 Pythia checkpoints test verified exposure with fold-local estimation; the complete primary estimates and controls follow.\par}
\vspace{4pt}
\begin{center}
\begin{minipage}{\columnwidth}
\centering
\scriptsize
\setlength{\tabcolsep}{3pt}
\begin{tabular}{lccc}
\toprule
\textbf{Model} & \textbf{LTS} & \textbf{L2} & \textbf{likelihood} \\
\midrule
Pythia-70M & 0.505 [0.492, 0.518] & 0.521 [0.502, 0.540] & 0.491 [0.476, 0.505] \\
Pythia-160M & 0.510 [0.497, 0.524] & 0.473 [0.459, 0.488] & 0.478 [0.469, 0.487] \\
Pythia-410M & 0.508 [0.495, 0.520] & 0.493 [0.477, 0.509] & 0.495 [0.481, 0.509] \\
Pythia-1B & 0.505 [0.492, 0.516] & 0.490 [0.474, 0.506] & 0.505 [0.489, 0.521] \\
Pythia-2.8B & 0.508 [0.496, 0.520] & 0.526 [0.507, 0.547] & 0.500 [0.485, 0.515] \\
\bottomrule
\end{tabular}
\vspace{4pt}
\begin{tabular}{lc}
\toprule
\textbf{Model-free method} & \textbf{AUC [95\% CI]} \\
\midrule
word TF--IDF & 0.491 [0.477, 0.504] \\
char TF--IDF & 0.488 [0.475, 0.500] \\
joint TF--IDF & 0.488 [0.475, 0.501] \\
metadata & 0.497 [0.483, 0.512] \\
MiniLM & 0.504 [0.489, 0.519] \\
\bottomrule
\end{tabular}
\captionof{table}{\textbf{Complete exact-checkpoint primary estimates.} Values are pooled out-of-fold ROC--AUC [cluster-bootstrap 95\% CI] on 3,994 probes/1,997 reciprocal groups, 20 split seeds, and 10,000 resamples. Every transform and hyperparameter is fitted inside the outer-training fold. Each of the five model-free methods is printed once because the gate verifies bitwise-identical estimates and intervals at every scale.}
\label{tab:checkmia-primary-full}
\end{minipage}
\end{center}

\begin{table}[!htbp]
\centering
\scriptsize
\setlength{\tabcolsep}{2.5pt}
\begin{tabular}{llccccc}
\toprule
\textbf{Model} & \textbf{Comparator} & \textbf{LTS} & \textbf{Comparator} & \textbf{LTS $-$ comparator} & \textbf{$p$} & \textbf{Holm $p$} \\
\midrule
Pythia-70M & L2 & 0.505 [0.492, 0.518] & 0.521 [0.502, 0.540] & -0.015 [-0.035, +0.005] & 0.169 & 0.634 \\
Pythia-70M & likelihood & 0.505 [0.492, 0.518] & 0.491 [0.476, 0.505] & +0.014 [-0.005, +0.034] & 0.158 & 0.634 \\
Pythia-70M & word TF--IDF & 0.505 [0.492, 0.518] & 0.491 [0.477, 0.504] & +0.015 [-0.004, +0.033] & 0.097 & 0.487 \\
Pythia-70M & char TF--IDF & 0.505 [0.492, 0.518] & 0.488 [0.475, 0.500] & +0.018 [-0.000, +0.036] & 0.042 & 0.293 \\
Pythia-70M & joint TF--IDF & 0.505 [0.492, 0.518] & 0.488 [0.475, 0.501] & +0.017 [-0.001, +0.035] & 0.052 & 0.310 \\
Pythia-70M & metadata & 0.505 [0.492, 0.518] & 0.497 [0.483, 0.512] & +0.008 [-0.011, +0.027] & 0.406 & 0.813 \\
Pythia-70M & MiniLM & 0.505 [0.492, 0.518] & 0.504 [0.489, 0.519] & +0.001 [-0.019, +0.021] & 0.914 & 0.914 \\
\midrule
Pythia-160M & L2 & 0.510 [0.497, 0.524] & 0.473 [0.459, 0.488] & +0.037 [+0.018, +0.056] & $<0.001$ & 0.001 \\
Pythia-160M & likelihood & 0.510 [0.497, 0.524] & 0.478 [0.469, 0.487] & +0.032 [+0.017, +0.048] & $<0.001$ & 0.001 \\
Pythia-160M & word TF--IDF & 0.510 [0.497, 0.524] & 0.491 [0.477, 0.504] & +0.020 [+0.001, +0.038] & 0.028 & 0.085 \\
Pythia-160M & char TF--IDF & 0.510 [0.497, 0.524] & 0.488 [0.475, 0.500] & +0.023 [+0.004, +0.041] & 0.012 & 0.059 \\
Pythia-160M & joint TF--IDF & 0.510 [0.497, 0.524] & 0.488 [0.475, 0.501] & +0.022 [+0.003, +0.040] & 0.015 & 0.059 \\
Pythia-160M & metadata & 0.510 [0.497, 0.524] & 0.497 [0.483, 0.512] & +0.013 [-0.006, +0.033] & 0.177 & 0.353 \\
Pythia-160M & MiniLM & 0.510 [0.497, 0.524] & 0.504 [0.489, 0.519] & +0.006 [-0.014, +0.026] & 0.544 & 0.544 \\
\midrule
Pythia-410M & L2 & 0.508 [0.495, 0.520] & 0.493 [0.477, 0.509] & +0.015 [-0.004, +0.034] & 0.117 & 0.469 \\
Pythia-410M & likelihood & 0.508 [0.495, 0.520] & 0.495 [0.481, 0.509] & +0.013 [-0.005, +0.031] & 0.135 & 0.469 \\
Pythia-410M & word TF--IDF & 0.508 [0.495, 0.520] & 0.491 [0.477, 0.504] & +0.017 [-0.000, +0.035] & 0.045 & 0.227 \\
Pythia-410M & char TF--IDF & 0.508 [0.495, 0.520] & 0.488 [0.475, 0.500] & +0.020 [+0.004, +0.037] & 0.016 & 0.111 \\
Pythia-410M & joint TF--IDF & 0.508 [0.495, 0.520] & 0.488 [0.475, 0.501] & +0.020 [+0.003, +0.037] & 0.021 & 0.126 \\
Pythia-410M & metadata & 0.508 [0.495, 0.520] & 0.497 [0.483, 0.512] & +0.011 [-0.008, +0.029] & 0.244 & 0.488 \\
Pythia-410M & MiniLM & 0.508 [0.495, 0.520] & 0.504 [0.489, 0.519] & +0.004 [-0.015, +0.022] & 0.725 & 0.725 \\
\midrule
Pythia-1B & L2 & 0.505 [0.492, 0.516] & 0.490 [0.474, 0.506] & +0.015 [-0.004, +0.033] & 0.141 & 0.562 \\
Pythia-1B & likelihood & 0.505 [0.492, 0.516] & 0.505 [0.489, 0.521] & -0.000 [-0.019, +0.018] & 0.974 & 1.000 \\
Pythia-1B & word TF--IDF & 0.505 [0.492, 0.516] & 0.491 [0.477, 0.504] & +0.014 [-0.004, +0.032] & 0.106 & 0.529 \\
Pythia-1B & char TF--IDF & 0.505 [0.492, 0.516] & 0.488 [0.475, 0.500] & +0.017 [-0.001, +0.034] & 0.047 & 0.331 \\
Pythia-1B & joint TF--IDF & 0.505 [0.492, 0.516] & 0.488 [0.475, 0.501] & +0.016 [-0.002, +0.034] & 0.054 & 0.331 \\
Pythia-1B & metadata & 0.505 [0.492, 0.516] & 0.497 [0.483, 0.512] & +0.007 [-0.011, +0.025] & 0.444 & 1.000 \\
Pythia-1B & MiniLM & 0.505 [0.492, 0.516] & 0.504 [0.489, 0.519] & +0.000 [-0.019, +0.019] & 0.979 & 1.000 \\
\midrule
Pythia-2.8B & L2 & 0.508 [0.496, 0.520] & 0.526 [0.507, 0.547] & -0.018 [-0.040, +0.003] & 0.104 & 0.418 \\
Pythia-2.8B & likelihood & 0.508 [0.496, 0.520] & 0.500 [0.485, 0.515] & +0.008 [-0.010, +0.025] & 0.361 & 0.723 \\
Pythia-2.8B & word TF--IDF & 0.508 [0.496, 0.520] & 0.491 [0.477, 0.504] & +0.018 [+0.000, +0.035] & 0.035 & 0.175 \\
Pythia-2.8B & char TF--IDF & 0.508 [0.496, 0.520] & 0.488 [0.475, 0.500] & +0.021 [+0.004, +0.037] & 0.012 & 0.083 \\
Pythia-2.8B & joint TF--IDF & 0.508 [0.496, 0.520] & 0.488 [0.475, 0.501] & +0.020 [+0.003, +0.037] & 0.017 & 0.100 \\
Pythia-2.8B & metadata & 0.508 [0.496, 0.520] & 0.497 [0.483, 0.512] & +0.011 [-0.007, +0.029] & 0.222 & 0.665 \\
Pythia-2.8B & MiniLM & 0.508 [0.496, 0.520] & 0.504 [0.489, 0.519] & +0.004 [-0.015, +0.023] & 0.701 & 0.723 \\
\bottomrule
\end{tabular}
\caption{\textbf{Complete fold-aligned primary comparisons.} Differences use identical examples, groups, folds, and split seeds; intervals use 10,000 paired group bootstraps and $p$-values use 10,000 paired group permutations with Holm correction within each model.}
\label{tab:checkmia-paired-full}
\end{table}

\begin{table}[!htbp]
\centering
\scriptsize
\setlength{\tabcolsep}{3pt}
\begin{tabular}{llccccc}
\toprule
\textbf{Model} & \textbf{Contrast} & \textbf{LTS} & \textbf{L2} & \textbf{LTS $-$ L2} & \textbf{$p$} & \textbf{Holm $p$} \\
\midrule
Pythia-70M & relevant $-$ no context & 0.505 [0.489, 0.521] & 0.512 [0.495, 0.529] & -0.007 [-0.027, +0.013] & 0.503 & 0.503 \\
Pythia-70M & relevant $-$ shuffled & 0.500 [0.485, 0.516] & 0.488 [0.473, 0.502] & +0.013 [-0.008, +0.033] & 0.230 & 0.230 \\
\midrule
Pythia-160M & relevant $-$ no context & 0.513 [0.496, 0.529] & 0.516 [0.500, 0.532] & -0.003 [-0.023, +0.016] & 0.717 & 0.717 \\
Pythia-160M & relevant $-$ shuffled & 0.499 [0.484, 0.515] & 0.502 [0.486, 0.517] & -0.002 [-0.025, +0.020] & 0.858 & 0.858 \\
\midrule
Pythia-410M & relevant $-$ no context & 0.495 [0.480, 0.510] & 0.491 [0.477, 0.506] & +0.004 [-0.015, +0.023] & 0.676 & 0.676 \\
Pythia-410M & relevant $-$ shuffled & 0.512 [0.496, 0.528] & 0.514 [0.497, 0.531] & -0.002 [-0.024, +0.019] & 0.830 & 0.830 \\
\midrule
Pythia-1B & relevant $-$ no context & 0.507 [0.491, 0.523] & 0.488 [0.473, 0.502] & +0.019 [-0.000, +0.039] & 0.056 & 0.056 \\
Pythia-1B & relevant $-$ shuffled & 0.481 [0.468, 0.494] & 0.507 [0.490, 0.524] & -0.026 [-0.048, -0.004] & 0.020 & 0.020 \\
\midrule
Pythia-2.8B & relevant $-$ no context & 0.496 [0.482, 0.511] & 0.515 [0.499, 0.531] & -0.019 [-0.038, +0.000] & 0.054 & 0.054 \\
Pythia-2.8B & relevant $-$ shuffled & 0.503 [0.489, 0.517] & 0.514 [0.497, 0.531] & -0.011 [-0.032, +0.010] & 0.304 & 0.304 \\
\bottomrule
\end{tabular}
\caption{\textbf{Complete exact-checkpoint structural controls.} Values are pooled out-of-fold ROC--AUC [95\% CI]. LTS--L2 differences use 10,000 paired group bootstraps and permutation tests over the same 3,994 probes/1,997 groups and 20 split seeds. Each prespecified model--condition LTS--L2 contrast is a singleton family, so its displayed Holm and raw $p$-values coincide.}
\label{tab:checkmia-structural-full}
\end{table}
% ===== End generated_verified_exposure_tables.tex =====

The discovery comparisons retain the earlier Cognitive Reality Monitoring
(CRM) labels, inspired by \citet{johnson1993reality}. They indicate access to
model computation: level 1 uses paired output-text distances with BGE-M3
\citep{bge-m3,reimers2019sbert}, level 2 uses output-head KL divergence, and
level 3 uses hidden states. These access levels differ from the exposure,
choice, and control questions in the main text. LTS is the hidden-state
measurement; its definition and intervention do not depend on a
psychological model of reality monitoring.

\begin{table}[!htbp]
\centering
\refstepcounter{section}
{\large\bfseries \thesection\quad Discovery Breadth Is Not Confirmatory
Attribution\par}
\vspace{2pt}
{\small The temporal-label WikiMIA screen can encode lexical and date
shortcuts, so the complete nine-model result is retained as discovery
evidence and is never reused as confirmatory source-choice evidence.\par}
\vspace{4pt}
\scriptsize
\setlength{\tabcolsep}{3.5pt}
\begin{tabular}{lccccc}
\toprule
Model & Best likelihood & LTS--LR [95\% CI] & LTS--XGB [95\% CI] &
Gain over likelihood & LTS dim. \\
\midrule
Llama-3.1-8B        & 0.565 & 0.778 [0.741, 0.838] & 0.707 [0.655, 0.759] & +0.213 & 12 \\
Llama-3.1-8B-Inst   & 0.575 & 0.708 [0.665, 0.760] & 0.627 [0.553, 0.687] & +0.133 & 12 \\
Mistral-7B-v0.3     & 0.575 & 0.869 [0.843, 0.890] & 0.815 [0.777, 0.856] & +0.294 & 12 \\
Mistral-7B-Inst     & 0.596 & 0.799 [0.744, 0.845] & 0.731 [0.680, 0.769] & +0.203 & 12 \\
Qwen2.5-7B          & 0.583 & 0.784 [0.765, 0.801] & 0.765 [0.739, 0.791] & +0.201 & 10 \\
Qwen2.5-7B-Inst     & 0.579 & 0.869 [0.837, 0.901] & 0.815 [0.777, 0.860] & +0.290 & 10 \\
Qwen2.5-14B         & 0.562 & 0.840 [0.793, 0.874] & 0.777 [0.756, 0.797] & +0.278 & 17 \\
Qwen2.5-14B-Inst    & 0.567 & 0.951 [0.941, 0.961] & 0.864 [0.842, 0.885] & +0.384 & 17 \\
Qwen2.5-32B-Inst    & 0.550 & 0.923 [0.881, 0.957] & 0.880 [0.815, 0.923] & +0.373 & 23 \\
\midrule
Mean                & 0.572 & 0.836 & 0.778 & +0.264 & -- \\
\bottomrule
\end{tabular}
\caption{\textbf{Complete original nine-model WikiMIA discovery screen.}
Best likelihood is the maximum of perplexity, zlib-normalized perplexity, and
Min-\(K\%\) probability.  Brackets are bootstrap intervals.  The temporal
label and shared-calibration protocol make this a breadth result, not verified
model-relative exposure.}
\label{tab:legacy-nine-model}
\end{table}

\begin{table}[!htbp]
\centering
\scriptsize
\setlength{\tabcolsep}{4pt}
\begin{tabular}{p{0.18\textwidth}p{0.19\textwidth}p{0.40\textwidth}p{0.12\textwidth}}
\toprule
Analysis & Scope & Complete retained result & Interpretation \\
\midrule
Model-free WikiMIA audit &
Same 250 rows &
Word TF--IDF 0.965 [0.945, 0.982]; word+character 0.962
[0.940, 0.980]; character 0.939 [0.912, 0.962]; length/style
0.645 [0.579, 0.710] &
Strong shortcut warning \\
Magnitude baseline &
Nine models &
Multilayer L2 mean AUC 0.812 versus LTS--LR 0.836; LTS matches or
exceeds L2 on 5/9 models &
Distinct, neither universal \\
Same-topic control &
Qwen-14B, Mistral-7B, Llama-8B &
LTS--LR 0.921, 0.842, 0.726; changes from random pairing are
\(-0.004,+0.020,-0.058\) &
Topic does not explain all signal \\
Prompt/label controls &
Same three models &
Four-template standard deviations 0.019, 0.009, 0.011; ten label
permutations return \(0.50\pm0.05\) &
Classifier and prompt checks pass \\
BookMIA &
Qwen-14B, Mistral-7B, Llama-8B &
Continuation LTS--LR 0.844, 0.967, 0.905; QA 0.980, 0.959, 0.969;
L2 0.683, 0.823, 0.745 &
External benchmark transfer \\
MIMIR negative control &
Pile-Wikipedia split &
LTS AUC range 0.48--0.55 &
No universal membership detector \\
Multi-task breadth &
Six model variants &
Continuation 0.788--0.969; summarization 0.815--0.964; QA
0.830--0.967 &
Task-dependent magnitude \\
\bottomrule
\end{tabular}
\caption{\textbf{Compact retention of the legacy robustness and boundary
analyses.}  These rows preserve positive, mixed, and negative outcomes.  The
MIMIR negative control (Duan et al.\ 2024; full citation in the main paper)
and the model-free WikiMIA audit prevent the discovery screen from being
interpreted as verified exposure or direct source use.}
\label{tab:legacy-boundaries}
\end{table}
% ===== End supplement_submission_content.tex =====
\end{document}